\pdfoutput=1
\documentclass{article}

\usepackage[nonatbib,preprint]{neurips_2025}

\usepackage[utf8]{inputenc} 
\usepackage[T1]{fontenc}    
\usepackage{hyperref}       
\usepackage{url}            
\usepackage{booktabs}       
\usepackage{amsfonts}       
\usepackage{nicefrac}       
\usepackage{microtype}      
\usepackage{xcolor}         
\usepackage{colortbl}

\definecolor{mygreen}{rgb}{0.1294,0.6980,0.6706}
\definecolor{myblue}{rgb}{0.2784,0.6784,1}

\usepackage{multirow}
\usepackage{algorithm}
\usepackage[noend]{algorithmic}
\renewcommand{\algorithmicrequire}{}  
\usepackage{setspace}
\usepackage{nicematrix}
\usepackage{amssymb}

\usepackage{subfig}

\title{AFCL: Analytic Federated Continual Learning for Spatio-Temporal Invariance of Non-IID Data}

\newcommand{\affiliationID}[1]{\textsuperscript{\rm{#1}}}

\author{%
    \hspace{-1.25em}%
    Jianheng Tang\affiliationID{1}\hspace{0.3em}%
    Huiping Zhuang\affiliationID{2}\hspace{0.3em}%
    Jingyu He\affiliationID{3}\hspace{0.3em}%
    Run He\affiliationID{2}\hspace{0.3em}%
    Jingchao Wang\affiliationID{1}\hspace{0.3em}%
    Kejia Fan\affiliationID{3}\hspace{0.3em}%
    Anfeng Liu\affiliationID{3}\\
    \textbf{
    \hspace{-1.25em}%
    Tian Wang\affiliationID{4}\hspace{0.3em}%
    Leye Wang\affiliationID{1}\hspace{0.3em}%
    Zhanxing Zhu\affiliationID{5} \hspace{0.3em}%
    Shanghang Zhang\affiliationID{1} \hspace{0.3em}%
    Houbing Song\affiliationID{6} \hspace{0.3em}%
    Yunhuai Liu\affiliationID{1}
    }\\
    \hspace{-1.6em}%
    \affiliationID{1}Peking University, China
    \hspace{0.5em}
    \affiliationID{2}South China University of Technology, China\\
    \hspace{-1.6em}%
    \affiliationID{3}Central South University, China
    \hspace{0.5em}
    \affiliationID{4}Beijing Normal University, China
    \\
    \hspace{-1.6em}%
    \affiliationID{5}University of Southampton, UK
    \hspace{0.5em}
    \affiliationID{6}University of Maryland, Baltimore County, USA\\
}

\begin{document}

\maketitle


\vspace{-0.25cm}
\begin{abstract}
\vspace{-0.05cm}
Federated Continual Learning (FCL) enables distributed clients to collaboratively train a global model from online task streams in dynamic real-world scenarios.
However, existing FCL methods face challenges of both spatial data heterogeneity among distributed clients and temporal data heterogeneity across online tasks.
Such data heterogeneity significantly degrades the model performance with severe spatial-temporal catastrophic forgetting of local and past knowledge.
In this paper, we identify that the root cause of this issue lies in the inherent vulnerability and sensitivity of gradients to non-IID data.
To fundamentally address this issue, we propose a gradient-free method, named Analytic Federated Continual Learning (AFCL), by deriving analytical (i.e., closed-form) solutions from frozen extracted features.
In local training, our AFCL enables single-epoch learning with only a lightweight forward-propagation process for each client.
In global aggregation, the server can recursively and efficiently update the global model with single-round aggregation.
Theoretical analyses validate that our AFCL achieves spatio-temporal invariance of non-IID data.
This ideal property implies that, regardless of how heterogeneous the data are distributed across local clients and online tasks, the aggregated model of our AFCL remains invariant and identical to that of centralized joint learning.
Extensive experiments show the consistent superiority of our AFCL over state-of-the-art baselines across various benchmark datasets and settings.

\end{abstract}    
\vspace{-0.2cm}
\section{Introduction}
\label{sec:intro}
\vspace{-0.05cm}

Federated Learning (FL) is a widely adopted paradigm that enables distributed clients to collaboratively train machine learning models while preserving the privacy of their respective data~\cite{FL-new-0, FL-new-1, FL-new-2}.
In traditional FL, the data heterogeneity (i.e., non-IID issue) across distributed clients represents one of the most critical challenges, leading to a degradation in the performance of the aggregated global model on local tasks~\cite{fl-new-3, fl-new-4, fl-new-5}.
To address this issue, many studies have been proposed and have made some progress~\cite{fl-new-6, fl-new-7, fl-new-8}, yet they are typically based on an unrealistic static assumption that the domains and categories of the training data of all clients remain unchanged~\cite{FCL-2, FCL-7, FCL-8-FedWeIT}.

\begin{figure}
    \centering
    \includegraphics[width=1.0\linewidth]{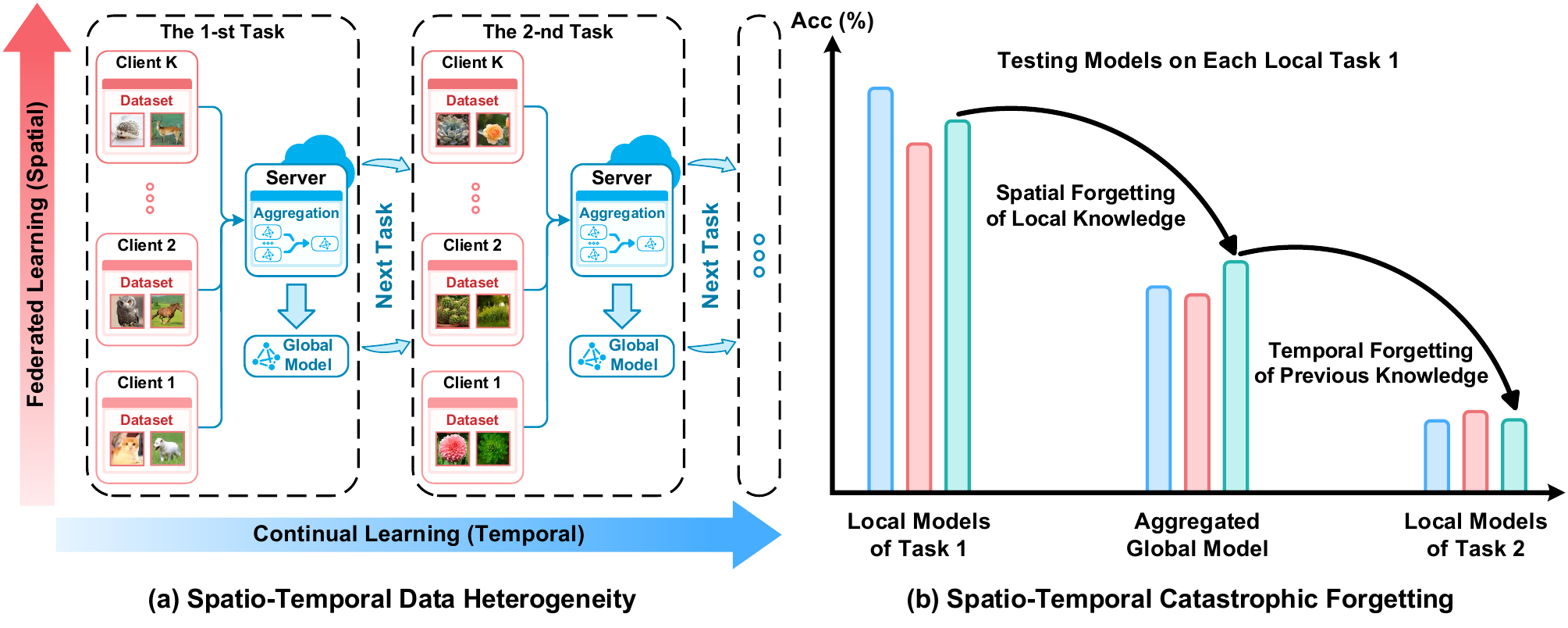}
    \caption{The spatial-temporal data heterogeneity and  catastrophic forgetting in FCL.}
    \label{fig:fig1}
    \vspace{-0.53cm}
\end{figure}

Federated Continual Learning (FCL) breaks the static limitations by enabling clients to continuously acquire new knowledge from online task streams~\cite{FCL-4, FCL-5, FCL-Prompt-0}.
While FCL expands the applicability of FL in dynamic real-world scenarios, it introduces a more challenging issue: spatio-temporal data heterogeneity~\cite{FCL-0, FCL-1, FCL-3}.
This heterogeneity arises not only across different clients (spatial level) but also within different tasks of the same client (temporal level), as shown in Figure~\ref{fig:fig1} (a).

The most direct negative impact of spatial-temporal data heterogeneity is spatial-temporal catastrophic forgetting~\cite{FCL-1, FCL-3}, as illustrated in Figure~\ref{fig:fig1} (b).
Catastrophic forgetting is originally a term from Continual Learning (CL), used to describe the phenomenon where deep models, after being trained on new data samples, tend to rapidly forget previously acquired knowledge~\cite{CIL_3_replay, ACIL_2}.
In FCL, the local models face temporal catastrophic forgetting as new data continue to arrive over time for both existing clients and incoming clients~\cite{FCL-11-FCIL, FCL-12}.
Meanwhile, the non-IID data across distributed clients can cause the aggregated global model to forget local knowledge of each client in a spatial sense~\cite{FCL-10-Lander, FCL-13-bad}.
In addition, as the local clients employ the aggregated global model as the basis to further learn the next task, the spatial and temporal forgetting tend to mutually influence each other, thereby making directly applying CL to FL infeasible~\cite{FCL-1, FCL-11-FCIL}.
We provide more detailed analyses in Appendix~\ref{subsection:appendix.non-iid}.

Spatio-temporal data heterogeneity can severely disrupt the stability of gradient-based updates to affect the model parameters~\cite{FCL-1, FCL-3}.
In this paper, we identify that the root cause of spatial-temporal catastrophic forgetting lies in the inherent vulnerability and sensitivity of gradients to non-IID data.
Specifically, as data distributions evolve over time and space, gradient-based updates tend to overwrite well-learned model parameters from previous tasks, consequently leading to spatio-temporal catastrophic forgetting.
Thus, the key insight of this work is to fundamentally address spatio-temporal catastrophic forgetting in FCL by avoiding the primary culprit of gradients.
Recent studies have shown the effectiveness of frozen pre-trained models in handling spatial-temporal heterogeneity~\cite{FCL-1, cil-ranpac, fl-pre-train-1, fl-pre-train-3}, as the frozen pre-trained models possess sufficient capacity to extract features without requiring gradient-based updates to their parameters~\cite{FCL-Prompt-0, FCL-3, fl-pre-train-2, CIL-pre-train}. 
These studies’ findings validate our insight and present a timely opportunity for our work.

To fundamentally address the issue of spatial-temporal data heterogeneity and bridge the gap in FCL, we propose a gradient-free method, named \underline{\textbf{A}}nalytic \underline{\textbf{F}}ederated \underline{\textbf{C}}ontinual \underline{\textbf{L}}earning (AFCL).
In our AFCL, each client shares a frozen pre-trained model as a powerful gradient-free feature extractor.
Then, we further construct a gradient-free classifier with the technique of the least
squares method, which can be trained by directly deriving analytical (i.e., closed-form) solutions from the frozen extracted features.
Moreover, we introduce a known-unknown class splitting mechanism to support the challenging class-continual scenario in FCL, where arbitrary class-level heterogeneity exists in the clients' online data increments.
The main contributions of this paper are summarized as follows:

\begin{itemize}
    \item Identifying the instability of gradients as the root cause of spatio-temporal catastrophic forgetting, we propose AFCL to fundamentally address this issue by avoiding gradients in training.
    To our knowledge, our AFCL represents the first gradient-free method in FCL.
    \item For the client side, we devise an analytic local training stage in our AFCL to enable single-epoch learning with only a lightweight forward-propagation process by leveraging the least squares method.
    For the server side, we propose a global recursive aggregating stage in our AFCL to recursively update the global model with single-round aggregation, which further avoids the high communication overhead introduced by multiple rounds.
    \item Theoretical analyses confirm that our AFCL can achieve spatio-temporal invariance of non-IID data. This is an ideal and rare property in FCL, implying that regardless of how heterogeneous the data are distributed across local clients and online tasks, the aggregated model of our AFCL remains invariant and identical to that of centralized joint learning.
    \item Extensive experiments show the consistent superiority of our AFCL over state-of-the-art baselines across a wide range of datasets and settings, with little to no performance degradation irrespective of the increasing degree of spatio-temporal data heterogeneity.
\end{itemize}

\section{Related Work}
\label{sec:related}



\subsection{Federated Continual Learning}
As a prevalent distributed learning paradigm, FL has made significant progress, particularly in tackling spatial data heterogeneity across distributed clients~\cite{fl-new-3, fl-new-4, fl-new-5}.
Unfortunately, traditional FL studies are typically based on an unrealistic static assumption that the domains and categories of training data across all clients remain unchanged, thus failing to handle more realistic and dynamic scenarios where each client continuously learns from its own evolving task stream~\cite{fl-new-6, fl-new-7, fl-new-8}.
In this context, by introducing the concept of CL to FL, FCL has emerged as a prevalent topic for distributed knowledge acquisition from the clients' online data without forgetting previous knowledge~\cite{FCL-2, FCL-7, FCL-8-FedWeIT}.

In a recent survey on FCL~\cite{FCL-0}, researchers highlight that the interaction between spatial and temporal heterogeneity has been largely overlooked in existing studies, leading to the unique challenge of spatio-temporal catastrophic forgetting~\cite{FCL-4, FCL-11-FCIL, FCL-10-Lander}.
To date, although a few studies have attempted to address data heterogeneity from the temporal-spatial perspective, none of them have truly recognized gradients as the root cause of forgetting, causing them to fail in fundamentally resolving spatio-temporal catastrophic forgetting~\cite{FCL-1, FCL-3}.
To the best of our knowledge, by proposing our gradient-free AFCL, we are the first to achieve the ideal property of spatio-temporal invariance in FCL.

\subsection{Analytic Learning}
Analytic learning is a representative gradient-free technique, originally widely developed to address gradient-related issues, e.g., vanishing and exploding gradients~\cite{ACIL_2,ACIL_1}.
Due to its flexible usage of matrix inversion, analytic learning is also referred to as pseudoinverse learning~\cite{AL_1, AL_new_0, AL_new_1}.
The radial basis network represents a typical example of shallow analytic learning, employing least squares estimation to train parameters after the kernel transformation in the first layer~\cite{AL_2}.
Moreover, by transforming nonlinear network learning into linear segments through least squares techniques, multilayer analytic learning has also found numerous applications~\cite{AL_3, AL_4, AL_CNN-2, AL_5}.

Nonetheless, earlier analytic learning techniques faced a severe memory challenge as they required processing the entire dataset simultaneously~\cite{AL_6}.
This limitation is mitigated by the block-wise recursive Moore-Penrose inverse, which effectively replaces joint learning with recursive learning~\cite{AL_6}.
Building on this advancement, the concept of analytic learning has achieved remarkable performances in various tasks~\cite{cil-ranpac, ACIL_1, AFL, AL_RL}, particularly in the field of CL~\cite{ACIL_new-0, ACIL_new-1, ACIL_new-2, ACIL_new-3}.
However, there still stands a significant gap in introducing analytic learning into FCL, due to the complex spatial-temporal data heterogeneity of FCL, where the domains and categories of training data may be arbitrarily mixed across both temporal and spatial dimensions.
Our AFCL aims to bridge the existing research gap, thereby fully leveraging the advantages of gradient-free analytic learning in FCL.

\section{Our Proposed AFCL}
\label{sec:method}

In FCL, the server needs to cope with two forms of heterogeneous incremental data: (1) new data continuously collected by existing clients, and (2) new data introduced by newly joined clients.
To uniformly and flexibly address the two types of temporal increments, our proposed AFCL adopts an asynchronous manner by treating all newly introduced data as if they were brought by newly joined clients.
Specifically, each batch of new data continuously collected by an existing client is treated as if it were introduced by a newly joined \textbf{\textit{virtual client}}.
Thus, the entire system can be modeled as consisting of a central server $\mathcal{S}$ and a set of $K$ (virtual) clients, denoted by $\mathcal{C} = \left\{ \mathcal{C}_1, \mathcal{C}_2, \cdots, \mathcal{C}_K \right\}$.
The $k$-th client's dataset is denoted as $\left. \mathcal{D}_{k} \right.\sim\left\{ {\mathcal{X}_{k},\mathcal{Y}_{k}} \right\}$, where $\mathcal{X}_{k} \in \mathbb{R}^{N_{k} \times \omega \times h \times c}$ (containing $N_{k}$ images with the shape of $\omega \times h \times c$) and $\mathcal{Y}_{k} \in \mathbb{R}^{N_{k}}$ (containing $N_{k}$ examples) correspond to the stacked input and label data for the $k$-th client, respectively. 
Moreover, we consider a challenging class-continual setting in FCL, where arbitrary class-level heterogeneity exists in the clients' online data increments.
In practice, the server is often unable to pre-determine the total number or specific types of classes that may appear during training.
Fortunately, each client can declare the set of classes contained in its current local data during registration, enabling the server to dynamically manage the class space (e.g., the one-hot encoding mapping).
Notably, our AFCL is not limited to image recognition, the most common task in the field of FCL~\cite{FCL-0, FCL-1, FCL-3}, and can be readily extended to other tasks.
For clarity, we summarize the descriptions of key notations in Table~\ref{table-nonation} of the appendix.

\clearpage
\subsection{Motivation and Overview}

\begin{figure*}[t]
    \centering
    \includegraphics[width=1.0\linewidth]{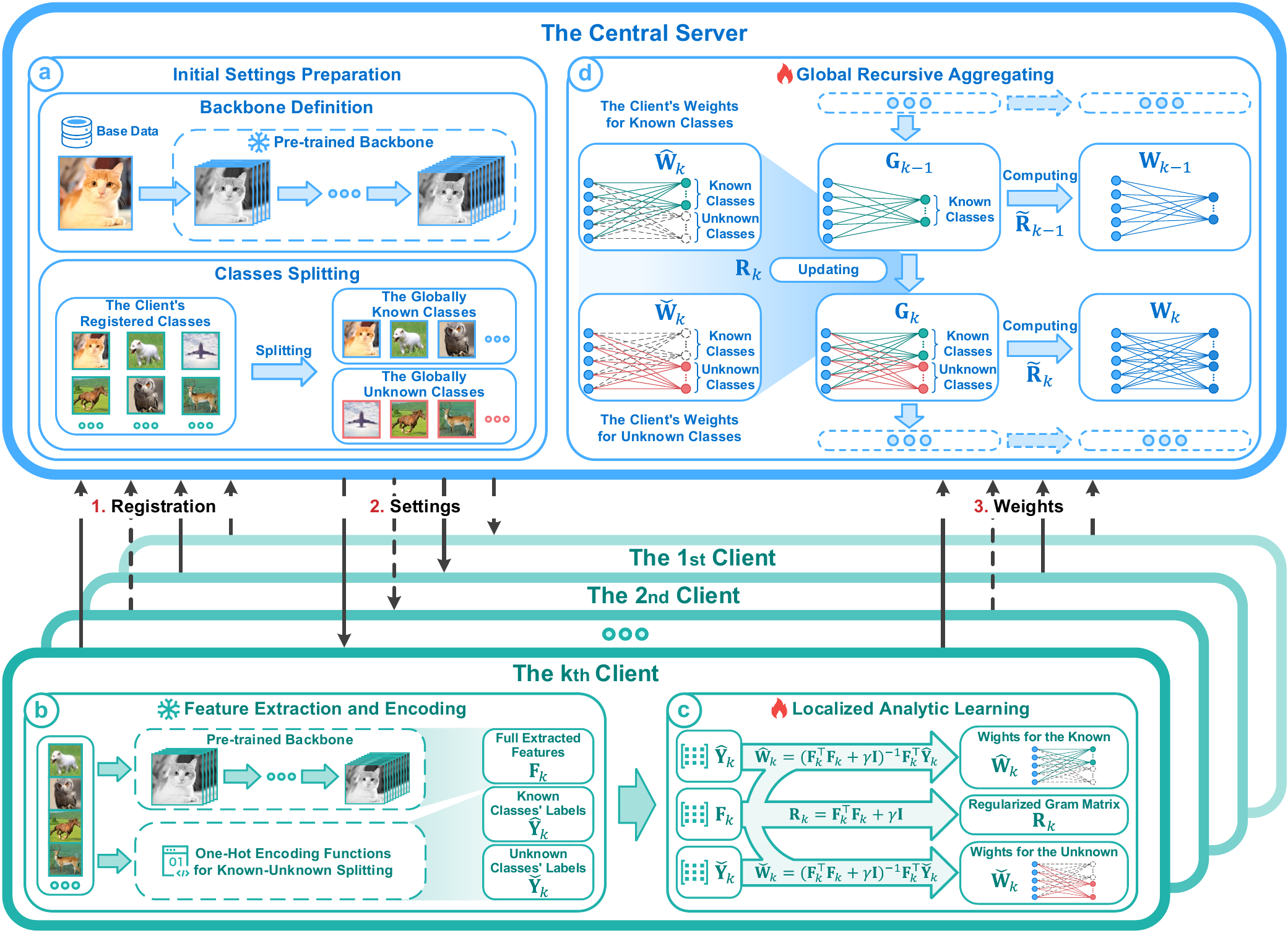}
    \caption{The framework of our proposed AFCL.}
    \label{fig:detail-framwork}
    \vspace{-0.15cm}
\end{figure*}

We provide a detailed analysis of spatio-temporal data heterogeneity in Appendix~\ref{subsection:appendix.non-iid}.
Based on it, we identify that the root cause of spatio-temporal catastrophic forgetting lies in the inherent vulnerability and sensitivity of gradients to non-IID data.
Motivated by this insight, our AFCL is designed to fundamentally address spatio-temporal catastrophic forgetting by avoiding the reliance on gradients.

The models within our AFCL mainly consist of two key components: a gradient-free feature extractor and a gradient-free analytic classifier.
Recent studies have demonstrated the effectiveness of the large pre-trained models in the fields of FL~\cite{fl-pre-train-1, fl-pre-train-3, fl-pre-train-2}, CL~\cite{cil-ranpac, CIL-pre-train, CIL-survey}, and FCL~\cite{FCL-Prompt-0,FCL-1, FCL-3}.
Thus, we use this opportunity by adopting a frozen pre-trained backbone as the gradient-free feature extractor with powerful representation capacity.
Next, we embed a linear analytic classifier after the feature extractor, with its final global parameters $\mathbf{W}_K$ optimized under the following objective:
\begin{equation}
\label{Motivation-1}
    \mathbf{W}_{K} 
    = \operatorname*{argmin}_{\mathbf{W}} \;{\left\|\mathbf{Y}_{1:K} -\mathbf{F}_{1:K} \mathbf{W} \right\|}_\mathrm{F}^2+\gamma{\left\| \mathbf{W} \right\|}_\mathrm{F}^2,
\end{equation}
where $\mathbf{F}_{1:K}$ is the stacked features, and $\mathbf{Y}_{1:K}$ is the stacked labels with global one-hot encoding, both assembled from the full datasets $\mathcal{D}_{1:K}$ of all $K$ clients.
The term $\gamma{\left\| \mathbf{W} \right\|}_\mathrm{F}^2$ is used for regularization, with adjustable $\gamma$ for flexibility.
Notably, we use Mean Squared Error (MSE) instead of cross-entropy as the loss function for the classification problem, to enable a closed-form least-squares solution that can be rapidly computed and continuously updated. 
The MSE loss has been shown to yield typically comparable performance to the cross-entropy loss~\cite{MSE-0,MSE-1}, and has been widely used in CL to avoid catastrophic forgetting~\cite{ACIL_2, cil-ranpac, ACIL_new-1, ACIL_new-2, ACIL_new-3}.
The framework of our AFCL is illustrated in Figure \ref{fig:detail-framwork}.

In our AFCL, each (virtual) client first registers with the server based on the set of classes contained in its current local data.
Then, the server dynamically manages the global class space and publishes the setting information, including the pre-trained backbone and the one-hot encoding functions with known-unknown class splitting.
After that, each client performs feature extraction and label encoding, followed by localized analytic computation within a single epoch through the least-squares method.
Finally, each client uploads the computation results to the server, and the server aggregates them to update the global model, which requires only a single round for each client.
Through this framework, our AFCL enables gradient-free distributed training for dynamic class-level data increments.

\subsection{Known-Unknown Class Splitting}
\label{subsection:3.3}

Here, we consider the class-continual setting of FCL, where the ongoing data stream may introduce new classes that have not been encountered by previous clients.
In practice, the server is often unable to pre-determine the total number or specific types of classes that may appear during training.
To tackle this challenge, we propose the known-unknown class split mechanism to enable dynamic adjustments of the one-hot mapping and model structure for the effective incorporation of new classes.
Specifically, each newly participating client needs to transmit the set of classes contained in its current local data to the server during registration.
The server then designates previously encountered classes as ``known classes'' and the newly introduced classes by this new client as ``unknown classes''.

For clarity, we denote the sets of known and unknown classes when the $k$-th client connects to the server as $\mathcal{\hat S}_{k}$ and $\mathcal{\check S}_{k}$, respectively. 
In addition, we denote the corresponding one-hot encoding functions of known and unknown classes as $\mathrm{Onehot}_{k}( \cdot )$ and $\mathrm{Onehot}^*_{k}( \cdot )$.
Thus, upon receiving the class information from the client, the server records the previously unknown classes $\mathcal{\check S}_{k}$ and generates a dedicated one-hot encoding function $\mathrm{Onehot}_{k}^*( \cdot )$ specifically for them in the $k$-th round.

Concurrently, the server updates the one-hot encoding function for the next round of known classes, $\mathrm{Onehot}_{k+1}( \cdot )$, based on the current $\mathrm{Onehot}_{k}( \cdot )$ and the newly generated $\mathrm{Onehot}_{k}^*( \cdot )$, i.e.,
\begin{equation}
\label{eq:subsection-3.3-1}
\mathrm{Onehot}_{k+1}(s) = 
\begin{cases}
\left[ \mathrm{Onehot}_{k}(s) \quad \quad \mathbf{0} \right], & \text{if } s \in \mathcal{\hat S}_{k}, \\
\left[ \mathbf{0} \quad \quad \mathrm{Onehot}_{k}^*(s) \right], & \text{if } s \in \mathcal{\check S}_{k}.
\end{cases}
\end{equation}

Subsequently, the server transmits the one-hot encoding functions $\mathrm{Onehot}_{k}( \cdot )$ and $\mathrm{Onehot}_{k}^*( \cdot )$ to the $k$-th client as part of the setting information.
In addition, the setting information also includes the pre-trained backbone network $\mathrm{Backbone}( \cdot , \mathbf{\Theta}_\text{B})$ to the $k$-th client, where $\mathbf{\Theta}_\text{B}$ represents the weights.

\subsection{Client-Side: Analytic Local Training}
\label{subsection:3.4}

After obtaining the setting information from the server, the client locally extracts features from its dataset inputs using the pre-trained backbone and maps the corresponding labels using the one-hot encoding functions, as shown in Figure~\ref{fig:detail-framwork} (b).
Specifically, for the $k$-th client, upon receiving the pre-trained backbone network $\mathrm{Backbone}(\cdot, \mathbf{\Theta}_\text{B})$, the client performs a forward propagation of the input of its raw local data $\mathcal{X}_k$ through the backbone to extract the feature matrix $\mathbf{F}_{k}$, i.e.,
\begin{equation}
\label{eq:subsection-3.4-1}
\mathbf{F}_{k} = \mathrm{Backbone}\left( {\mathcal{X}_{k},\mathbf{\Theta}_\text{B}} \right).
\end{equation}
Here, $\mathbf{F}_{k} \in \mathbb{R}^{N_{k} \times l_{e}}$ represents the feature matrix for the $k$-th client, with $N_{k}$ and $l_{e}$ denoting the sample size and embedding length, respectively.
Meanwhile, the client also employs the obtained one-hot encoding functions $\mathrm{Onehot}_{k}( \cdot )$ and $\mathrm{Onehot}_{k}^*( \cdot )$ to map labels from its local data $\mathcal{Y}_k$, i.e.,
\begin{equation}
\label{eq:subsection-3.4-2}
\mathbf{\hat Y}_k = \mathrm{Onehot}_{k} ( \mathcal{Y}_{k} ),\quad
\mathbf{\check Y}_k = \mathrm{Onehot}_{k}^* ( \mathcal{Y}_{k} ).
\end{equation}
Here, $\mathbf{\hat Y}_k \in \mathbb{R}^{N_{k} \times d_{k-1}}$ and $\mathbf{\check Y}_k \in \mathbb{R}^{N_{k} \times (d_{k} - d_{k-1})}$ represent the label matrices for known and unknown classes, respectively, where $d_{k}$ and $d_{k-1}$ represent the total number of classes registered by the server across the datasets from the first $k$ clients and the first $(k-1)$ clients, respectively.

Upon obtaining $\mathbf{F}_k$, $\mathbf{\hat{Y}}_k$, and $\mathbf{\check{Y}}_k$, the client subsequently performs localized analytic computation, as shown in Figure~\ref{fig:detail-framwork} (c).
Since the client's dataset may contain previously unseen classes for the server, structural variations may occur in the aggregated global model.
To address this issue, in our AFCL, each client needs to train two separate analytical models $\hat{\mathbf{W}}_k \in \mathbb{R}^{l_{e} \times d_{k-1}}$ and $\check{\mathbf{W}}_k \in \mathbb{R}^{l_{e} \times (d_{k} - d_{k-1})}$ for the known and unknown classes, respectively.
Specifically, the training of these two local models is reformulated into two linear regression problems, which can be analytically solved through the least-squares method, as presented in \eqref{eq:subsection-3.4-6} and \eqref{eq:subsection-3.4-7}, with derivations provided in Section~\ref{subsection:3.6}.
\begin{equation}
\label{eq:subsection-3.4-6}
  \hat{\mathbf{W}}_k=\underset{\mathbf{{W}}}{\operatorname*{argmin}} \; \|\mathbf{\hat Y}_k-\mathbf{F}_k\mathbf{W}\|_\mathrm{F}^2+\mathbf{\gamma} \| \mathbf{W} \|_{\mathrm{F}}^2
  =(\mathbf{F}_k^\top\mathbf{F}_k+\gamma\mathbf{I})^{-1}\mathbf{F}_k^\top\mathbf{\hat Y}_k,
\end{equation}
\begin{equation}
\label{eq:subsection-3.4-7}
  \check{\mathbf{W}}_k=\underset{\mathbf{W}}{\operatorname*{argmin}} \; \|\mathbf{\check Y}_k-\mathbf{F}_k\mathbf{W}\|_\mathrm{F}^2+\mathbf{\gamma} \| \mathbf{W} \|_{\mathrm{F}}^2
  =(\mathbf{F}_k^\top\mathbf{F}_k+\gamma\mathbf{I})^{-1}\mathbf{F}_k^\top\mathbf{\check Y}_k.
\end{equation}
Additionally, we further define the \textit{Regularized Gram Matrix} for the $k$-th client as $\mathbf{R}_k \in \mathbb{R}^{l_{e} \times l_{e}}$, which can be computed by \eqref{eq:subsection-3.4-8}.
\begin{equation}
\label{eq:subsection-3.4-8}
  \mathbf{R}_k = \mathbf{F}_k^\top\mathbf{F}_k+\gamma\mathbf{I}.
\end{equation}
The \textit{Regularized Gram Matrix} is subsequently utilized to assist the server in performing global model aggregation.
Upon finishing all the analytical computations as mentioned above, the $k$-th client then uploads its two trained local models (i.e., $\mathbf{\hat W}_k$ and $\mathbf{\check W}_k$) and the \textit{Regularized Gram Matrix} (i.e., $\mathbf{R}_k$), all of which cannot be used to infer the client's local data, in terms of both inputs and labels.

\subsection{Server-Side: Global Recursive Aggregating}
\label{subsection:3.5}
After the $k$-th client completes the analytic local training, the server subsequently performs global recursive aggregating based on the information received from the client (i.e., ${\mathbf{\hat W}}_k$, ${\mathbf{\check W}}_k$, and $\mathbf{R}_k$). Specifically, the server adopts a recursive approach to update the global model with only a single-round aggregation for each client, thereby avoiding the multi-round communication overhead.

Considering the class-continual setting of FCL, the newly joined (virtual) clients may dynamically introduce new classes that the server has never encountered.
The dynamic introduction of previously unknown classes can alter the structure of the analytic classifier, changing its output dimensionality from $d_{k-1}$ to $d_k$.
To enable the global model to flexibly accommodate the structural increments caused by previously unknown classes and simultaneously adjust the parameters of the existing structure based on new data from known classes, we design the \textit{Global Knowledge Matrix} for the $k$-th round as $\mathbf{G}_k \in \mathbb{R}^{l_{e} \times d_{k}}$, which captures the accumulated global knowledge from the beginning up to the $k$-th round.
Specifically, our designed \textit{Global Knowledge Matrix} is recursively updated by the central server at each round with ${\mathbf{\hat W}}_k$, ${\mathbf{\check W}}_k$, and $\mathbf{R}_k$, as follows:
\begin{equation}
\label{eq:subsection-3.5-2:GKM update}
\mathbf{G}_{k} = \begin{bmatrix} \underbrace{\mathbf{A}_{k}  \mathbf{G}_{k-1} + \mathbf{B}_k  \mathbf{\hat W}_k}_{Knowledge \; for \; Known \; Classes} & \underbrace{\mathbf{B}_k  \mathbf{\check W}_k}_{Knowledge \; for \;Unknown \;Classes} \end{bmatrix}, k\in \left(1,K \right],
\end{equation}
where
\begin{equation}
\label{eq:subsection-3.5-3}
\begin{cases}
\mathbf{A}_{k} = \mathbf{I} - (\mathbf{\tilde R}_{k-1})^{-1} \mathbf{R}_k (\mathbf{I} - (\mathbf{\tilde R}_{k})^{-1} \mathbf{R}_k), & \mathbf{\tilde R}_{k} = \sum_{i=1}^k \mathbf{R}_i , \\
\mathbf{B}_{k} = \mathbf{I} - (\mathbf{R}_k)^{-1} \mathbf{\tilde R}_{k-1} (\mathbf{I} - (\mathbf{\tilde R}_{k})^{-1} \mathbf{\tilde R}_{k-1}),  & \mathbf{R}_i = \mathbf{F}_i^\top \mathbf{F}_i +\gamma\mathbf{I}.
\end{cases}
\end{equation}

In the first round, as all the classes are previously unknown, there is no knowledge for the known classes, and $\mathbf{G}_1 = \mathbf{\check W}_1$ is a special case.
The recursive formula \eqref{eq:subsection-3.5-2:GKM update} for $\mathbf{G}_k$ expresses the knowledge update in an explicit and interpretable manner. Specifically, the former term in \eqref{eq:subsection-3.5-2:GKM update} represents the knowledge for known classes, which is a combination of the previously stored knowledge $\mathbf{A}_{k} \mathbf{G}_{k-1}$ and the newly acquired knowledge $\mathbf{B}_k \mathbf{\hat W}_k$. 
The latter term $\mathbf{B}_k \mathbf{\check W}_k$, which is \textit{concatenated into} $\mathbf{G}_k$, reflects the acquired knowledge for the previously unknown classes in the $k$-th round.

Notably, the server does not need to store all historical \textit{Regularized Gram Matrices}, as $\mathbf{\tilde R}_k$ can also be updated recursively.
As the clients continually join, the \textit{Global Knowledge Matrix} is dynamically updated to accommodate the incremental data, while preserving the knowledge and memory of all previously learned data.
Through the recursive computation for $\mathbf{G}_k$, we can further derive its closed-form general expression, with details in \textbf{Lemma 2} of Appendix~\ref{subsection:appendix.2}.
The explicit and closed-form formulation of the \textit{Global Knowledge Matrix} provides the key basis for our AFCL to achieve spatio-temporal invariance against non-IID data, as shown in Section~\ref{subsection:3.6}.

The \textit{Global Knowledge Matrix} is a dedicated unit for storing globally obtained knowledge within our AFCL, rather than the final global model itself.
After updating the \textit{Global Knowledge Matrix} $\mathbf{G}_{k}$, the server needs to further compute the corresponding global analytical model $\mathbf{W}_{k}$ by \eqref{eq:subsection-3.5-4}.
\begin{equation}
\label{eq:subsection-3.5-4}
  \begin{aligned}
\mathbf{W}_k=[\mathbf{\tilde R}_{k} - (k-1)\gamma \mathbf{I}]^{-1} \mathbf{\tilde R}_{k} \mathbf{G}_{k}.
\end{aligned}
\end{equation}
In this way, the server continuously and recursively executes the above process until all the $K$ clients' local models have been aggregated into the global model, as shown in Figure~\ref{fig:detail-framwork} (d).
Notably, as shown in Section \ref{subsection:3.6}, the final global model $\mathbf{W}_K$ obtained by our AFCL is equivalent to empirical risk minimization in \eqref{Motivation-1} with the full datasets $\mathcal{D}_{1:k}$ from all $K$ clients, ideally achieving \textit{spatio-temporal invariance for non-IID data} and \textit{order invariance across local clients}.
For clarity, we provide the detailed procedures of our AFCL in \textbf{Algorithm \ref{alg:AFCL}} of the appendix.

It is worth noting that, with simple operations, the training of each client can support asynchronous execution in our AFCL.
Specifically, during the registration phase, the server can accept multiple clients and determine the setting of the known-unknown class splitting sequentially according to the registration IDs.
Since the clients may vary in computational capacity and data volume, the order in which they submit their weights may not align with their assigned IDs.
Taking this into consideration, the server can wait for all $K$ clients to submit their data before performing batch aggregations.

\subsection{Theoretical Analyses}
\label{subsection:3.6}

In this subsection, we theoretically analyze the validity of our AFCL, and introduce its ideal properties.
First of all, we derive the analytical solution for the linear regression problem, as follows.

\noindent\textbf{Lemma 1:} For an optimization problem of the following form:
\begin{equation}
    \label{eq:theorem-1-1}
    \operatorname*{argmin}_{\mathbf{W}} \; 
    {\left\|\mathbf{Y} -\mathbf{F} \mathbf{W} \right\|}_\mathrm{F}^2+\gamma{\left\| \mathbf{W} \right\|}_\mathrm{F}^2,
\end{equation}
there exists a closed-form solution given by:
\begin{equation}
    \label{eq:theorem-1-2}
    \mathbf{W}=
    {({\mathbf{F}}^\top\mathbf{F}+\gamma\mathbf{I})}^{-1}{\mathbf{F}}^\top\mathbf{Y}.
\end{equation}
\textbf{\textit{Proof.}} See Appendix \ref{subsection:appendix.1}.


According to \textbf{Lemma 1}, we can verify that the local analytical models for the known and unknown classes of the $k$-th client (i.e., $\mathbf{\hat W}_k$ and $\mathbf{\check W}_k$) have closed-form solutions, as specified in \eqref{eq:subsection-3.4-6} and \eqref{eq:subsection-3.4-7}.
In addition, we can derive that the global optimization objective with empirical risk minimization over the full datasets $\mathcal{D}_{1:K}$ from all $K$ clients in \eqref{Motivation-1}, admits the following closed-form solution:
\begin{equation}
    \label{eq:theorem-1-full}
    \mathbf{W}_{K} = (\mathbf{F}_{1:K}^\top\mathbf{F}_{1:K} + \gamma \mathbf{I})^{-1} ( \mathbf{F}_{1:K}^\top\mathbf{Y}_{1:K} ).
\end{equation}
For \eqref{eq:theorem-1-full}, by setting $K$ to $k$, we can further obtain a more generalized closed-form expression of the optimal global analytical model $\mathbf{W}_{k}$ for the first $k$ clients, as given below.
\begin{equation}
    \label{eq:theorem-1-3}
    \mathbf{W}_{k} = (\mathbf{F}_{1:k}^\top\mathbf{F}_{1:k} + \gamma \mathbf{I})^{-1} ( \mathbf{F}_{1:k}^\top\mathbf{Y}_{1:k} ).
\end{equation}

Then, we show that the global analytical model obtained by \eqref{eq:subsection-3.5-4} is exactly equivalent to the optimal solution \eqref{eq:theorem-1-3} of empirical risk minimization over the full datasets of the first $k$ clients, as follows.

\noindent\textbf{Theorem 1:} 
The computation results for $\mathbf{W}_k$ by \eqref{eq:subsection-3.5-4} is exactly equivalent to \eqref{eq:theorem-1-3}, which corresponds to the optimal solution (i.e., the centralized joint learning) of empirical risk minimization over the full datasets $\mathcal{D}_{1:k}$ from the first $k$ clients.

\noindent \textbf{\textit{Proof.}} See Appendix \ref{subsection:appendix.2}.

Based on \textbf{Theorem 1}, we can further derive two ideal properties of our AFCL: \textit{spatio-temporal invariance for non-IID data}, and \textit{order invariance across local clients}, as follows.

\noindent\textbf{Theorem 2:} The final global model obtained by our AFCL is independent of the spatio-temporal data heterogeneity and client registration order, being identical to that obtained by centralized joint learning over the full dataset $\mathcal{D}_{1:K}$ from all $K$ clients.

\noindent \textbf{\textit{Proof.}} See Appendix \ref{subsection:appendix.3}.

Finally, we also analyze the efficiency of our AFCL on computation and communication, which is provided in Appendix \ref{subsection:appendix.4} for interested readers.

\section{Experiments}

\subsection{Experiment Setting}
\label{sec:Experimental Setup}

\textbf{Datasets \& Settings.} 
To comprehensively evaluate the performance of our proposed AFCL, we conduct extensive experiments on three benchmark datasets: CIFAR-100 \cite{dataset_1}, Tiny-ImageNet \cite{dataset_2}, and ImageNet-R \cite{dataset_ImageNet-R}.
Existing methods typically partition the training process into multiple tasks, each involving collaborative model training among multiple clients.
To align with existing methods, we adopt the well-established Si-Blurry setting \cite{Si-Blurry} to partition each dataset into $5$ or $10$ tasks to simulate varying temporal data heterogeneity.
Subsequently, we allocate each task’s dataset among $5$ clients using the Dirichlet distribution \cite{Dirichlet} with the concentration parameter $\alpha \in \{0.1, 0.2, 0.5, 1.0\}$ to simulate varying spatial data heterogeneity.
The partitioning details are provided in Appendix \ref{app:Dataset Setting}.

\textbf{Baselines \& Metrics.} 
In our experiments, we compare our AFCL against state-of-the-art baselines, encompassing both replay-based and replay-free baselines. 
Specifically, the replay-based baselines include TARGET \cite{FCL-7}, FedIcaRL \cite{iCaRL}, and FedCBC \cite{FCL-3}, while the replay-free baselines include Finetune~\cite{fedavg}, FedLwF \cite{LwF}, FedEwc \cite{Ewc}, and FedMGP \cite{FCL-2}.
For a fair comparison, we adopt ResNet-18 \cite{backbone} as the unified backbone across all methods. 
Additionally, we select average accuracy, forgetting, stability, and plasticity as the performance metrics and cumulative runtime as the efficiency metrics.
Additional experimental details and metric descriptions are provided in Appendix \ref{app:Model Training}.

\subsection{Experimental Results}
\begin{table*}[t]
    \centering
    \renewcommand{\arraystretch}{1.3}
    \caption{
    Performance comparison of average accuracy among our AFCL and other baselines.
    The parameters $T$ and $\alpha$ are designed to control the levels of temporal and spatial heterogeneity in FCL.
    The \textbf{bold} and \underline{underlined} results indicate the best and second-best performance, respectively.
    }
    \label{table:performance}
    \resizebox{\textwidth}{!}{
        \begin{NiceTabular}{ l c c c c c c c c c >{\columncolor{mygreen!15}} c >{\columncolor{myblue!15}} c} 
        \toprule
        \textbf{Dataset} & \textbf{Task} & \textbf{Setting} & \textbf{Finetune} & \textbf{FedLwF} & \textbf{TARGET} & \textbf{FedEwc} & \textbf{FedIcaRL} & \textbf{FedMGP} & \textbf{FedCBC} & \textbf{AFCL} & \textbf{Improve}  \\
        \cline{1-12}
        \multirow{4}{*}{CIFAR-100}  &  \multirow{2}{*}{$T=5$}  &  $\alpha = 0.1$  & $4.96$ & $8.37$ & $11.09$ & $12.64$ & $4.05$ & $23.87$ & $\underline{43.34}$ & $\textbf{58.56}$ & $35.12 \%$ \\
        & & $\alpha = 0.2$  & $6.35$ & $11.11$ & $25.80$ & $14.28$ & $5.52$ & $26.75$ & $\underline{44.93}$ & $\textbf{58.56}$ & $30.34 \%$ \\
        \cline{2-12}
        & \multirow{2}{*}{$T=10$}  & $\alpha = 0.1$  & $4.15$ & $4.18$ & $8.21$ & $7.92$ & $3.64$ & $15.21$ & $\underline{27.32}$ & $\textbf{58.56}$ & $114.35 \%$ \\
        & & $\alpha = 0.2$  & $4.98$ & $5.63$ & $10.94$ & $11.48$ & $3.93$ & $16.17$ & $\underline{29.02}$ & $\textbf{58.56}$ & $101.79 \%$ \\

        \cline{1-12}
        \multirow{4}{*}{Tiny-ImageNet}  & \multirow{2}{*}{$T=5$}  & $\alpha = 0.1$  & $1.85$ & $3.13$ & $4.67$ & $3.34$ & $2.05$ & $12.53$ & $\underline{31.13}$ & $\textbf{54.67}$ & $75.62 \%$ \\
        & & $\alpha = 0.2$  & $2.24$ & $3.62$ & $7.46$ & $4.56$ & $2.41$ & $14.82$ & $\underline{33.87}$ & $\textbf{54.67}$ & $61.41 \%$ \\
        \cline{2-12}
        &   \multirow{2}{*}{$T=10$} & $\alpha = 0.1$  & $1.32$ & $2.70$ & $2.32$ & $2.63$ & $1.84$ & $9.60$ & $\underline{22.48}$ & $\textbf{54.67}$ & $143.19 \%$ \\
        & & $\alpha = 0.2$  & $1.47$ & $2.21$ & $2.48$ & $3.39$ & $1.86$ & $10.10$ & $\underline{24.62}$ & $\textbf{54.67}$ & $122.06 \%$ \\

        \cline{1-12}
        \multirow{4}{*}{ImageNet-R} & \multirow{2}{*}{$T=5$} &  $\alpha = 0.5$  & $1.41$ & $3.13$ &  $3.95$ & $4.27$ & $2.03$ & $18.23$ & $\underline{23.51}$ & $\textbf{38.22}$ & $62.57 \%$ \\
        & & $\alpha = 1.0$  & $1.73$ & $3.19$ & $4.93$ & $4.71$ & $2.22$ & $18.69$ & $\underline{23.79}$ & $\textbf{38.22}$ & $60.66 \%$  \\
        \cline{2-12}
        &  \multirow{2}{*}{$T=10$} & $\alpha = 0.5$  & $1.47$ & $2.17$ & $0.96$ & $3.21$ & $1.57$ & $8.75$ & $\underline{15.72}$ & $\textbf{38.22}$ & $143.13 \%$ \\
        & & $\alpha = 1.0$  & $1.65$ & $2.48$ & $1.02$ & $3.30$ & $1.78$ & $9.33$ & $\underline{14.14}$ & $\textbf{38.22}$ & $170.30 \%$\\
        \bottomrule
        \end{NiceTabular}
     }
\end{table*}

\begin{figure*}[t] 
\centering 
\begin{minipage}[t]{0.329\linewidth}
		\centering
		\subfloat[CIFAR-100]{ 
              \centering  
               \label{fig:Temporal-Invariance-CIFAR-100}
               \includegraphics[width=1.0\textwidth]{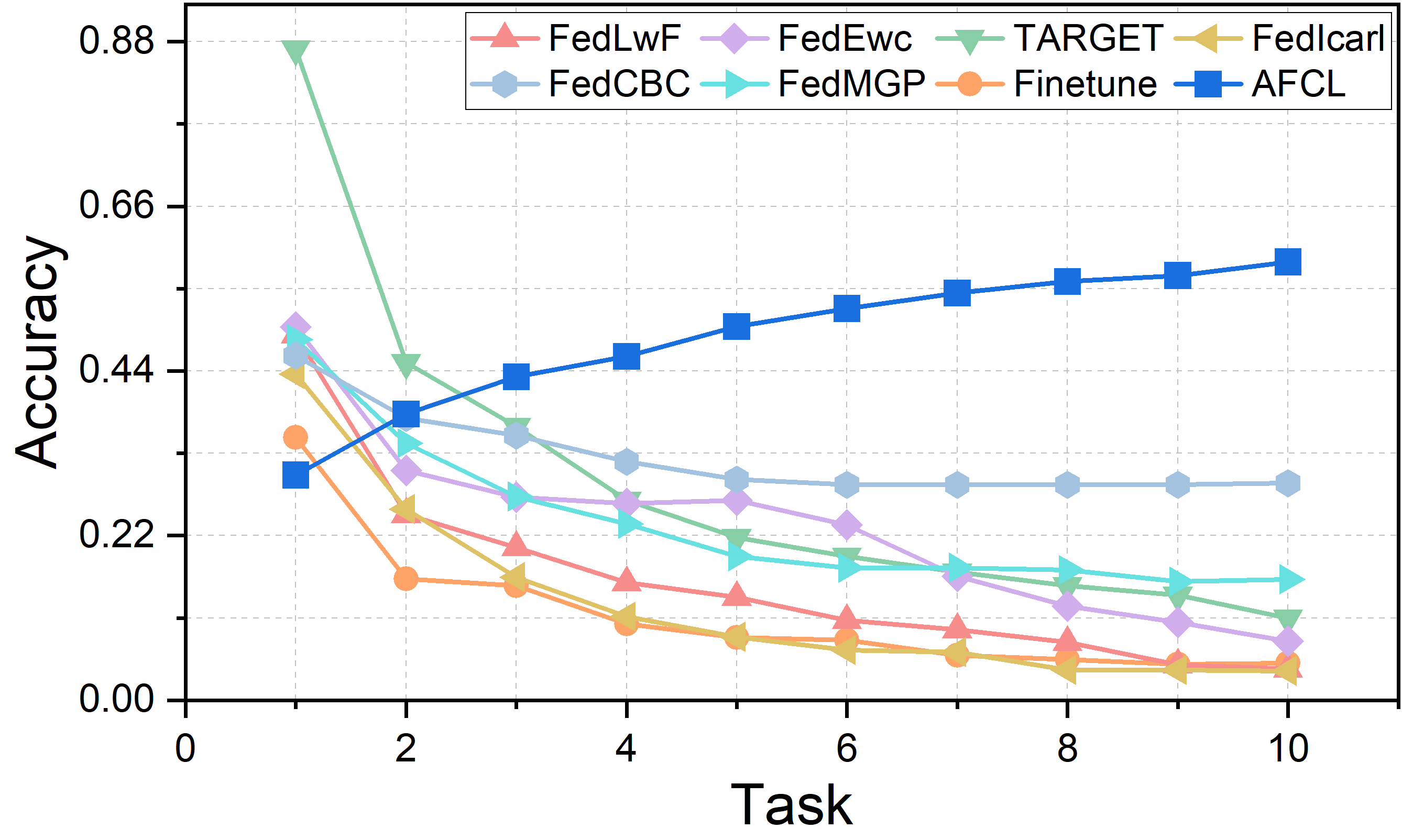}
             } 
	\end{minipage}
	\hfill 
	\begin{minipage}[t]{0.329\linewidth}
		\centering
		\subfloat[Tiny-ImageNet]{ 
              \centering  
               \label{fig:Temporal-Invariance-Tiny-ImageNet-free}
               \includegraphics[width=1.0\textwidth]{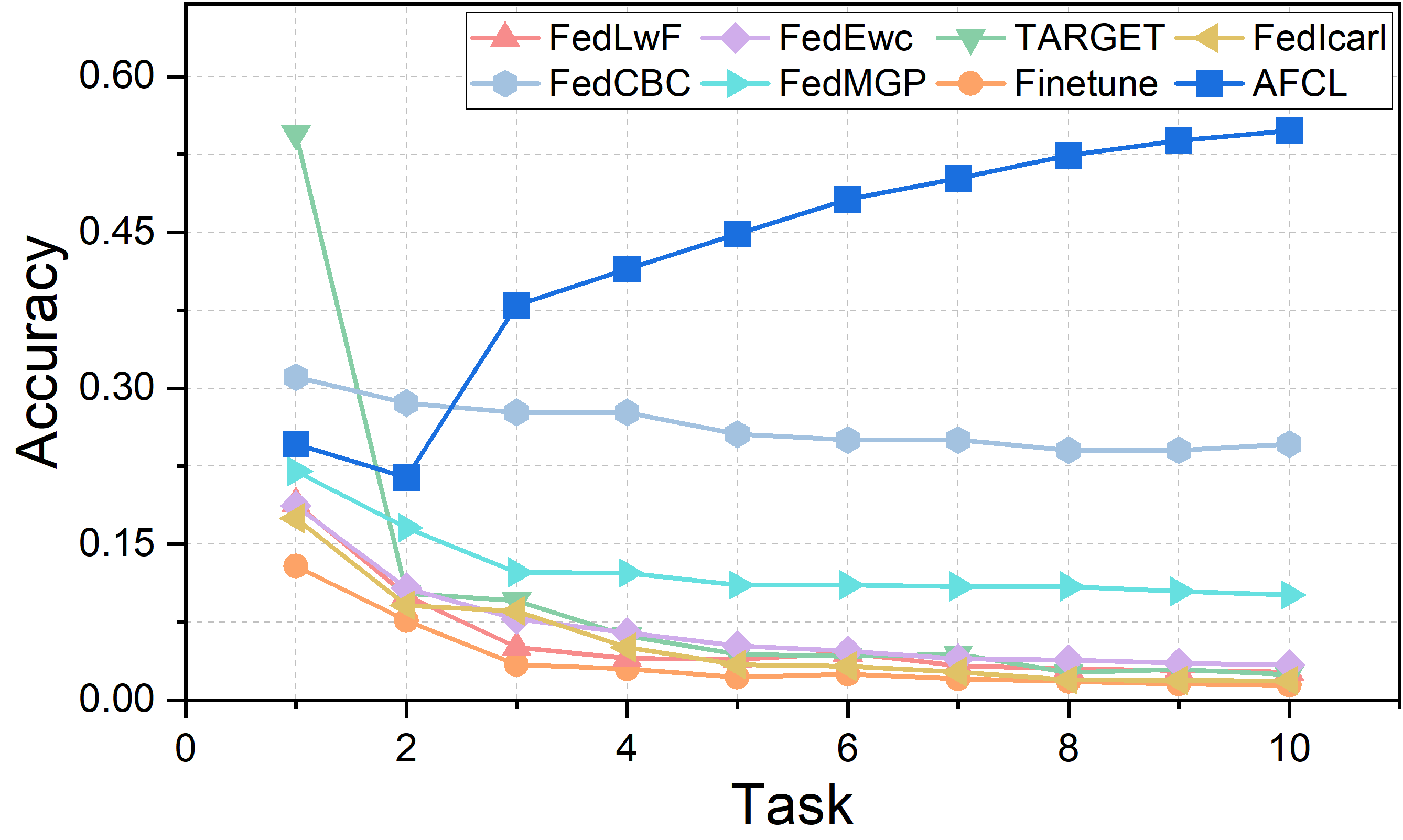}
             } 
	\end{minipage}
        \hfill 
	\begin{minipage}[t]{0.329\linewidth}
		\centering
		\subfloat[ImageNet-R]{ 
              \centering  
               \label{fig:Temporal-Invariance-ImageNet-R}
               \includegraphics[width=1.0\textwidth]{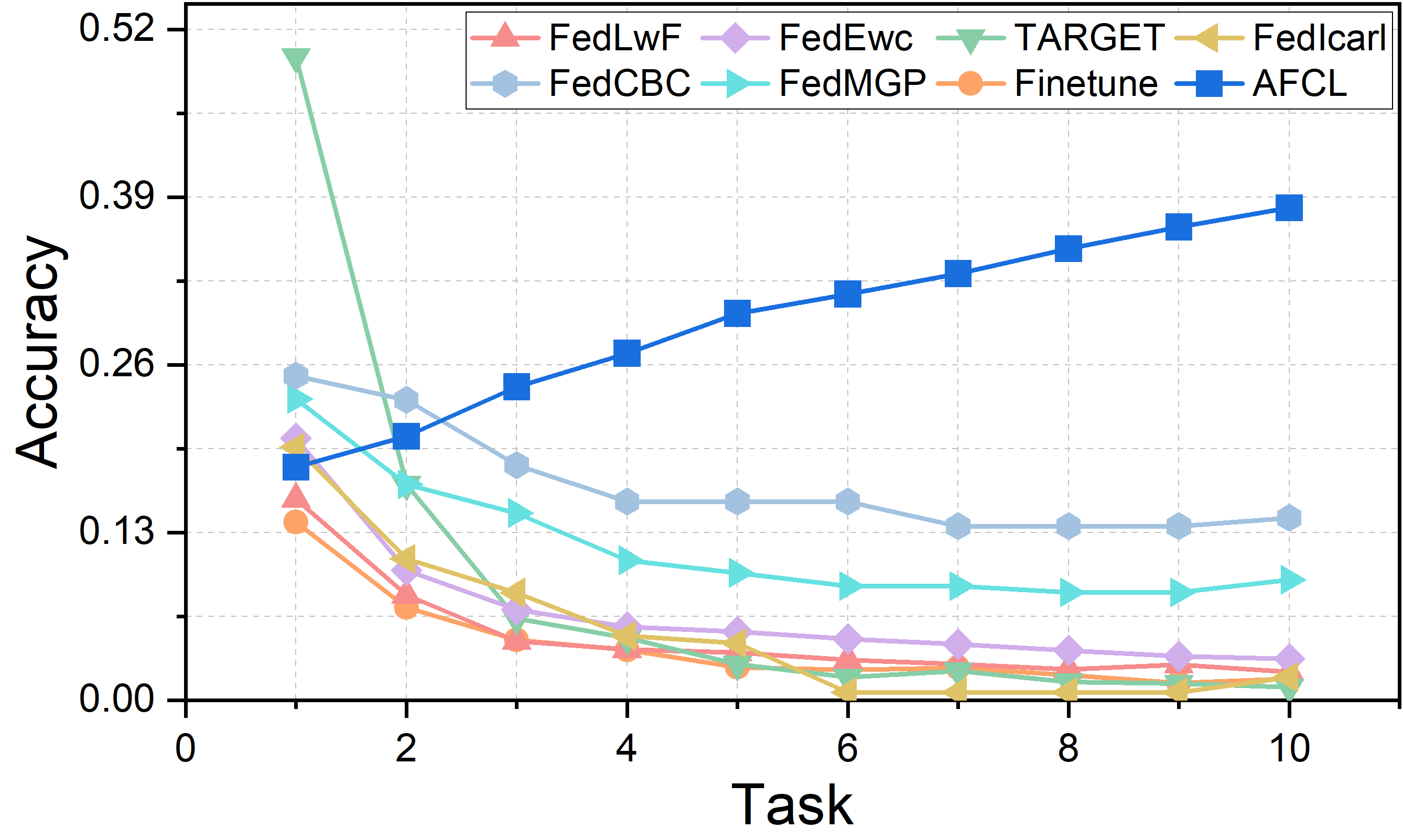}
             } 
	\end{minipage}
 \caption{Average accuracy of AFCL and the baselines among different tasks.}  
 \label{fig:accuracy}
\begin{minipage}[t]{0.329\linewidth}
		\centering
		\subfloat[CIFAR-100]{ 
              \centering  
               \label{fig:Temporal-Invariance-CIFAR-100}
               \includegraphics[width=1.0\textwidth]{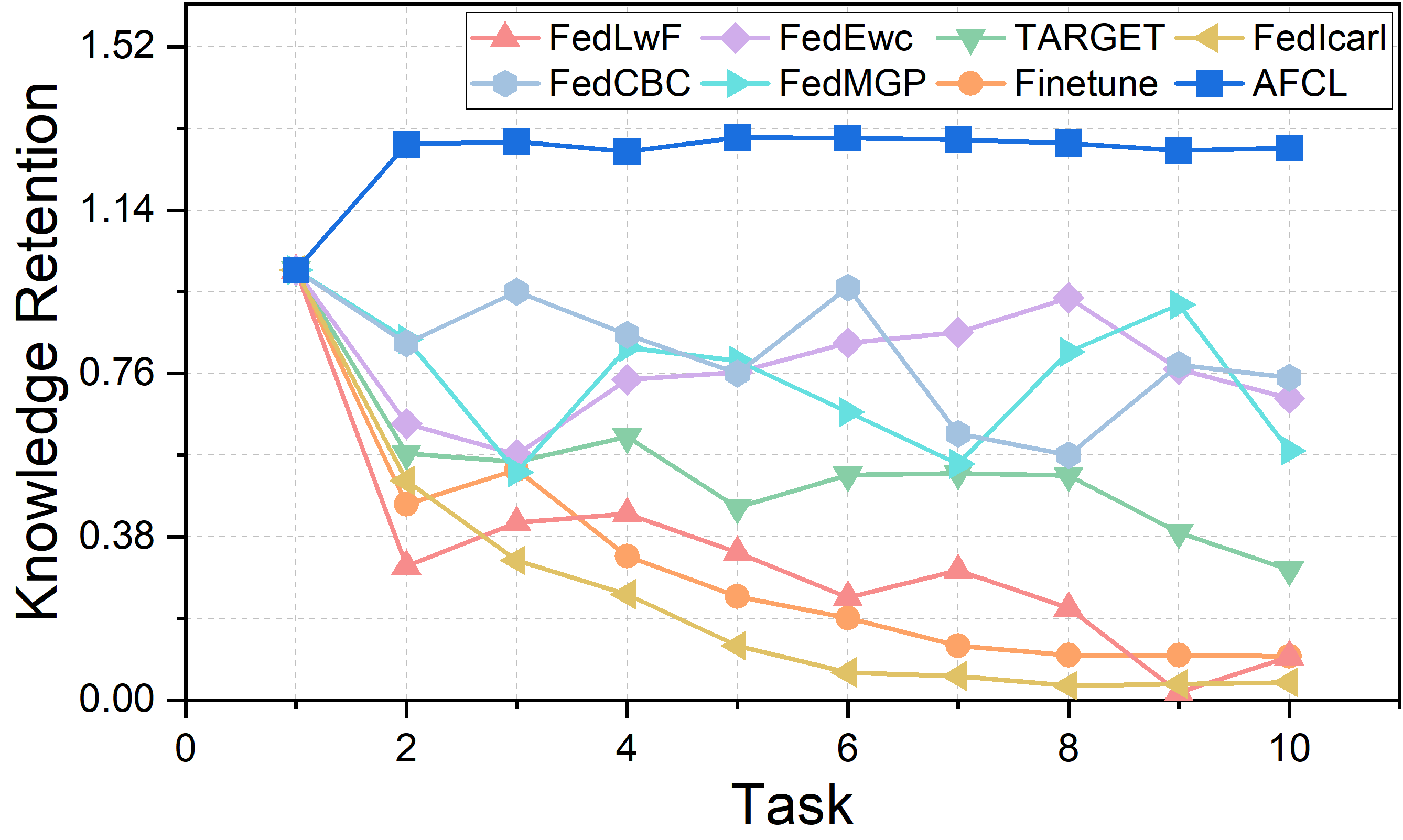}
             } 
	\end{minipage}
	\hfill 
	\begin{minipage}[t]{0.329\linewidth}
		\centering
		\subfloat[Tiny-ImageNet]{ 
              \centering  
               \label{fig:Temporal-Invariance-Tiny-ImageNet-free}
               \includegraphics[width=1.0\textwidth]{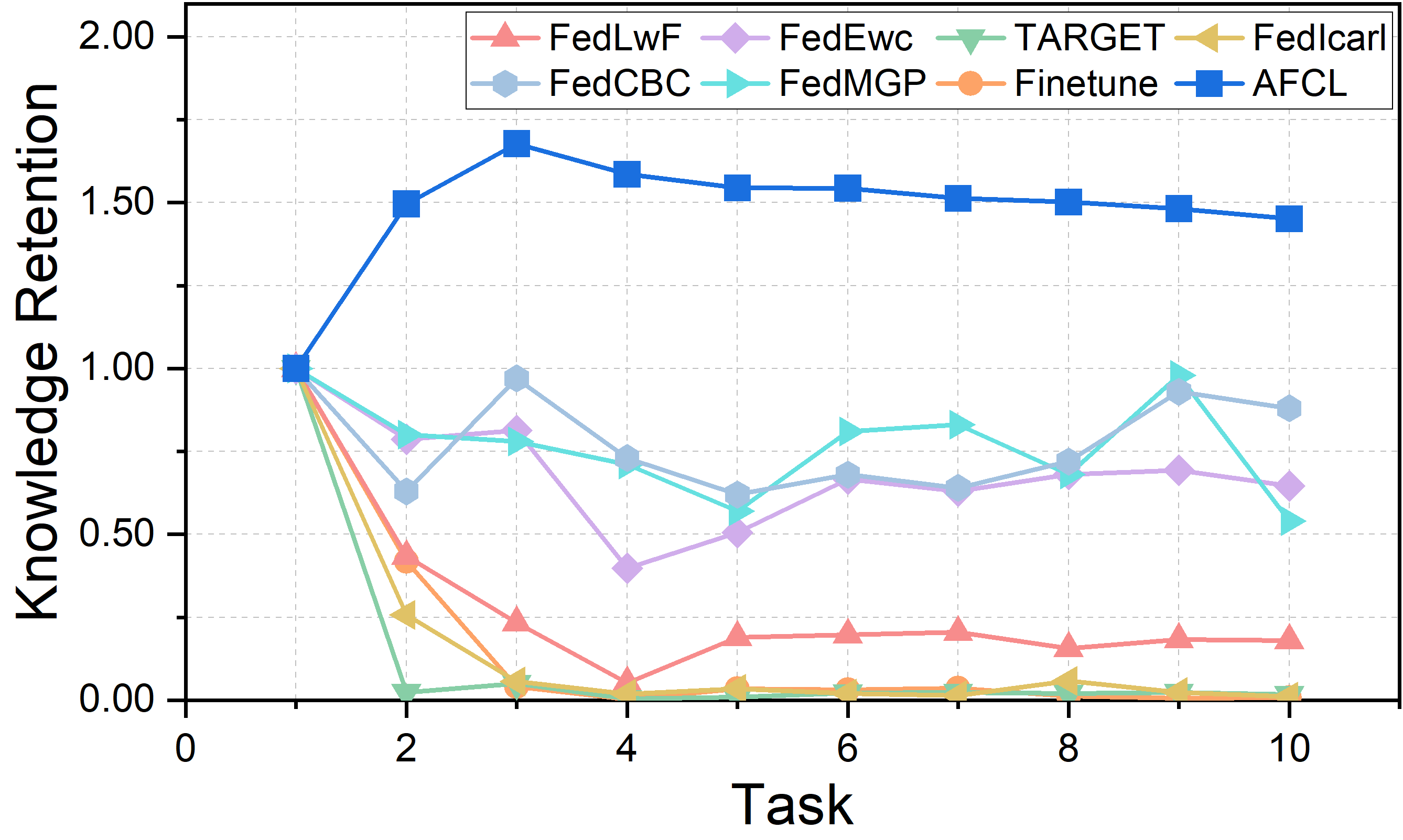}
             } 
	\end{minipage}
        \hfill 
	\begin{minipage}[t]{0.329\linewidth}
		\centering
		\subfloat[ImageNet-R]{ 
              \centering  
               \label{fig:Temporal-Invariance-ImageNet-R}
               \includegraphics[width=1.0\textwidth]{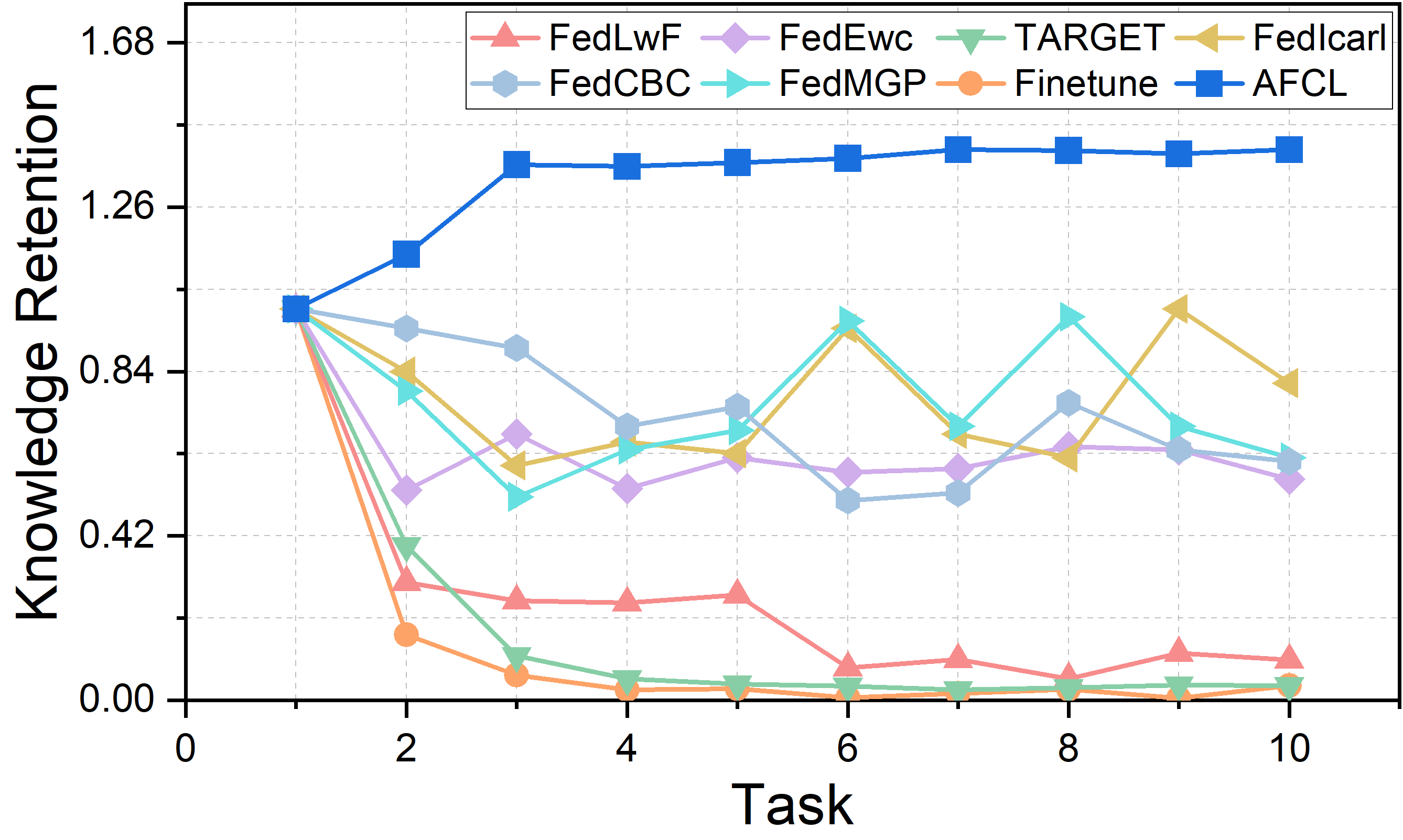}
             } 
	\end{minipage}
 \caption{Average knowledge retention of AFCL and the baselines among different tasks.}  
 \label{fig:forggeting}
    \vspace{-0.2cm}
\end{figure*}

\textbf{Spatio-Temporal Invariance.} 
First, we compare the performance of our AFCL with baselines under various spatio-temporal heterogeneity settings in Table \ref{table:performance}.
Our AFCL consistently achieves the best performance across all evaluated settings, outperforming the state-of-the-art method, FedCBC, by an average margin of up to $93.38\%$.
Notably, as temporal heterogeneity increases (i.e., $T$ rises from $5$ to $10$) or spatial heterogeneity intensifies (i.e., $\alpha$ decreases from $0.2$ to $0.1$ on CIFAR-100 and Tiny-ImageNet, and from $1.0$ to $0.5$ on ImageNet-R), the performance of all baselines exhibits a significant decline.
In particular, under the most challenging setting with high heterogeneity in both temporal and spatial dimensions (i.e., $T=10$ and $\alpha=0.1/0.5$), all baselines suffer severe performance collapse, with accuracy all dropping below $28\%$.
Conversely, benefiting from its desirable property of spatio-temporal invariance, our AFCL maintains stable and superior performance across all settings, with accuracy consistently achieving at least $38.22\%$ and reaching up to $58.56\%$.
Moreover, it is foreseeable that if spatio-temporal data heterogeneity in FCL further increases in practice, the advantage of our AFCL will be even more pronounced, thanks to its invariance to data heterogeneity.

\textbf{Spatio-Temporal Forgetting.}
Second, to further explore how model performance evolves as tasks progress, we examine the variations in average accuracy and knowledge retention across all methods, as shown in Figures~\ref{fig:accuracy} and~\ref{fig:forggeting}.
Specifically, the baselines suffer from severe spatio-temporal forgetting, resulting in a marked decrease in average accuracy and knowledge retention as tasks progress.
In contrast, as our AFCL continually learns from new tasks, its average accuracy shows an initial increase followed by stabilization, while maintaining the highest knowledge retention throughout the entire process.
Notably, the knowledge retention of our AFCL consistently exceeds 1.00, indicating its continual acquisition of new generalizable knowledge from incoming data.
In contrast, the knowledge retention of all baselines remains below 1.00 due to spatio-temporal forgetting.
These results provide strong evidence that our proposed AFCL effectively addresses the issue of spatio-temporal forgetting in FCL through a gradient-free method, fundamentally distinguishing it from existing baselines.

\textbf{Stability-Plasticity Analysis.}
Third, to more thoroughly investigate how different methods retain previous knowledge and acquire new knowledge, we compare them on stability (i.e., the ability to retain previous task knowledge) and plasticity (i.e., the ability to acquire new task knowledge) in Tables \ref{table:Stability} of Appendix~\ref{app:Stability-Plasticity}. 
These results highlight that the poor balance between stability and plasticity is an inherent limitation of existing gradient-based baselines, leading to the phenomenon of spatio-temporal catastrophic forgetting.
In contrast, our AFCL effectively overcomes this limitation by achieving both high plasticity and high stability in a gradient-free manner.

\textbf{Efficiency Analysis.}
Then, we also report the runtime overhead of our AFCL and the other baselines, as shown in Table \ref{table:Efficiency} of Appendix~\ref{app:Efficiency}.
Specifically, due to the baselines' reliance on gradient-based updates, they require each client to perform multiple epochs of training with costly back-propagation in every aggregation round.
Moreover, these aggregation rounds must be repeated across tasks to ensure convergence on the server side.
In contrast, our AFCL requires only a single epoch of lightweight forward-propagation on the client side in a gradient-free manner.
On the server side, our AFCL needs only a single-round aggregation for each client, thereby significantly reducing the communication cost associated with multiple rounds.
As a result, our AFCL shows remarkable priority in efficiency, reducing the total runtime by up to 90.92\% compared with the fastest baseline (i.e., Finetune) and by 97.43\% compared with the best-performing baseline (i.e., FedCBC).

\textbf{Sensitivity Analysis.}
Finally, we analyze the sensitivity of our AFCL to its only hyperparameter, the regularization term $\gamma$, as presented in Table \ref{table:sensitivity}.
We can observe that when $\gamma > 0$, the performance of our AFCL remains relatively stable across a wide range of parameter settings.
It indicates that our AFCL can achieve satisfactory performance without the need for extensive hyperparameter tuning.

\begin{table*}[t]
    \centering
    \renewcommand{\arraystretch}{1.3}
    \caption{Average accuracy of our AFCL with various values of the regularization term $\gamma$.}
    \label{table:sensitivity}
     \resizebox{1.0\textwidth}{!}{
        \begin{NiceTabular}{ l c c c c c c c c}
        \toprule
        \textbf{Dataset} & $\gamma = 0$ & $\gamma = 10^{-3}$ & $\gamma = 10^{-2}$ & $\gamma = 10^{-1}$ & $\gamma = 10^{0}$ & $\gamma = 10^{1}$ & $\gamma = 10^{2}$ & $\gamma = 10^{3}$ \\
        \cline{1-9}
        CIFAR-100 & $58.56$ & $58.56$ & $58.55$ & $58.54$ & $58.51$ & $58.15$ & $55.77$ & $51.18$ \\
        \cline{1-9}
        Tiny-ImageNet & $54.67$ & $54.65$ & $54.65$ & $54.64$ & $54.67$ & $54.23$ & $54.05$ & $48.32$ \\
        \cline{1-9} 
        ImageNet-R & $38.22$ & $38.21$ & $38.20$ & $38.22$ & $38.17$ & $38.05$ & $36.79$ & $31.22$ \\
        \bottomrule
        \end{NiceTabular}
        }
    \vspace{-0.335cm}
\end{table*}

\section{Conclusion and Discussion}
\label{sec:conclusion}

In this paper, we identify the instability of gradients as the root cause of spatio-temporal catastrophic forgetting in FCL and propose our AFCL to fundamentally address this issue by eliminating gradients during training. To the best of our knowledge, our AFCL is the first gradient-free method in FCL. Specifically, our AFCL leverages a frozen pre-trained feature extractor and develops a linear analytic classifier for gradient-free learning. 
To handle dynamic class-level data increments, we introduce a known-unknown class splitting mechanism, embedding the knowledge within a closed-formulation in an explicit and interpretable manner.
Theoretical analyses confirm two ideal properties of our AFCL, namely spatio-temporal invariance for non-IID data and order invariance across local clients.
Extensive experiments have shown the consistent superiority of AFCL over state-of-the-art baselines.

The major limitation of our AFCL is its reliance on a well-pretrained backbone as a frozen feature extractor.
In the current era of large AI models, this is not considered overly restrictive, and many studies in FCL adopt similar approaches. 
In particular, the pre-trained backbone can be readily obtained from a publicly available open-source model or trained centrally on open-sourced datasets, without introducing data privacy risks.
Nevertheless, this minor limitation could motivate further exploration into an adjustable backbone to enhance the capability of feature extraction.
Meanwhile, for the sake of simplicity, the analytic classifier AFCL only captures the linear relationship between the extracted features and the output labels. 
Although our current AFCL has already achieved superior performance, there is still promising potential for further improving its classification capability by incorporating techniques such as random projections, kernel methods, and ensemble learning.

\clearpage



\medskip


\bibliographystyle{unsrt}


\clearpage
\setcounter{page}{1}


\appendix

\begin{table*}[t]
    \centering
    \renewcommand{\arraystretch}{1.25}
    \caption{Description of important notations.
    }
    \label{table-nonation}
        \begin{NiceTabular}{ p{0.15\textwidth} p{0.79\textwidth} }
        \toprule
        \textbf{Notations} & \textbf{Description} \\
        \midrule
        $\mathcal{X}_{k}$ & The $k$-th client's raw input dataset.\\
        $\mathcal{Y}_{k}$ & The $k$-th client's raw label dataset.\\
        $\mathbf{F}_{k}$ & The $k$-th client's full feature matrix.\\
        $\mathbf{\hat Y}_{k}$ & The $k$-th client's label matrix for known classes.\\
        $\mathbf{\check Y}_{k}$ & The $k$-th client's label matrix for unknown classes.\\
        $\mathbf{\hat W}_{k}$ & The weights of the $k$-th client's local model for known classes.\\
        $\mathbf{\check W}_{k}$ & The weights of the $k$-th client's local model for unknown classes.\\
        $\mathbf{R}_{k}$ & The $k$-th client's \textit{Regularized Gram Matrix}. \\
        $\mathbf{\widetilde R}_{k}$ & The sum of the \textit{Regularized Gram Matrices} for first $k$ clients. \\
        $\mathbf{G}_{k}$ & The \textit{Global Knowledge Matrix} in the $k$-th round. \\
        $\mathbf{W}_{k}$ & The weights of the global model in the $k$-th round. \\
        \bottomrule
        \end{NiceTabular}
\end{table*}

\begin{algorithm}[t]
\setstretch{1.25}
\renewcommand{\algorithmicrequire}{\textbf{Input:}}
\renewcommand{\algorithmicensure}{\textbf{Output:}}
\renewcommand{\algorithmicreturn}{\textbf{Return:}}
\begin{algorithmic}[1]
\caption{Analytic Federated Continual Learning}
\label{alg:AFCL}
\REQUIRE The clients' local datasets $\{\mathcal{D}_1 , \mathcal{D}_2, \cdots , \mathcal{D}_K \}$.
\ENSURE The server's final global model $\mathbf{W}_K$.
\STATE The server initializes $\mathbf{W}_0 = \mathbf{0}$, $\mathbf{R}_0 = \mathbf{0}$, $\mathbf{G}_{0} = \mathbf{0}$;
\STATE The server determines $\mathrm{Backbone} \left( \cdot ,\mathbf{\Theta}_\text{B} \right)$ and $\mathrm{Tanh} \left( \cdot ,\mathbf{\Theta}_\text{T} \right)$;
\FOR {each round $k$ with the $k$-th client}
\STATE The client registers with the server and sends its class information;
\STATE The server performs classes splitting and sends the result to the client;
\STATE The client performs feature extraction to obtain feature matrix $\mathbf{F}_k$ by \eqref{eq:subsection-3.4-1};
\STATE The client performs label encoding to obtain label matrices $\mathbf{\hat Y}_k$ and $\mathbf{\check Y}_k$ by \eqref{eq:subsection-3.4-2};
\STATE The client calculates its local models $\mathbf{\hat W}_k$ and $\mathbf{\check W}_k$ by \eqref{eq:subsection-3.4-6} and \eqref{eq:subsection-3.4-7}; 
\STATE The client calculates its \textit{Regularized Gram Matrix} $\mathbf{R}_k$ by \eqref{eq:subsection-3.4-8};
\STATE The client sends $\mathbf{\hat W}_k$, $\mathbf{\check W}_k$, and $\mathbf{R}_k$ to the server;
\STATE The server updates the \textit{Global Knowledge Matrix} to obtain $\mathbf{G}_k$ by \eqref{eq:subsection-3.5-2:GKM update} and \eqref{eq:subsection-3.5-3};
\ENDFOR
\STATE The server calculates the global model $\mathbf{W}_K$ by \eqref{eq:subsection-3.5-4};
\RETURN The server's final global model $\mathbf{W}_K$.
\end{algorithmic}
\end{algorithm}

\begin{figure}[h]
    \centering
    \includegraphics[width=0.90\linewidth]{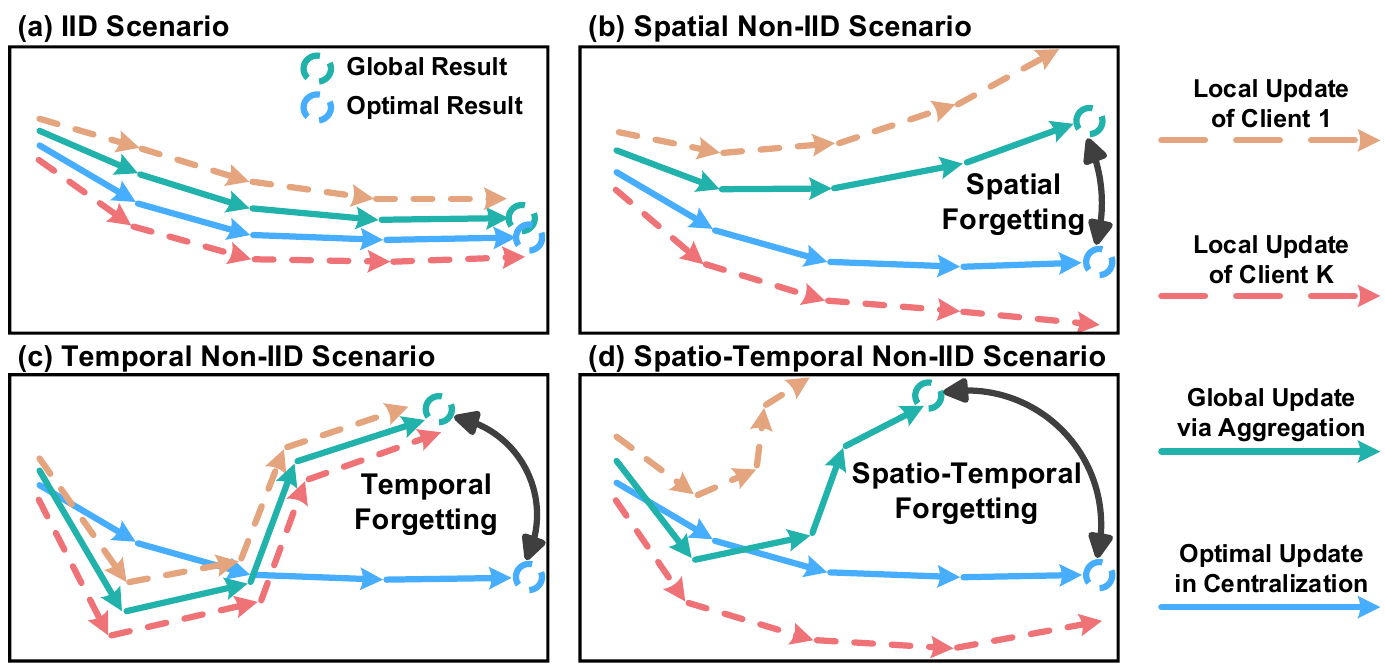}
    \caption{The gradient update perspective of catastrophic forgetting with non-IID data in FCL.}
    \label{fig:Impact-gradient}
    \vspace{-0.2cm}
\end{figure}

\newpage
\section{Appendix for Analysis on  Spatio-Temporal Data Heterogeneity} 
\label{subsection:appendix.non-iid}

As illustrated in Figure~\ref{fig:fig1} (a), in FCL, data heterogeneity occurs along both spatial and temporal dimensions.
Spatial heterogeneity refers to the fact that each client’s data is typically non-IID.
For example, different specialized hospitals may have data on different types of diseases.
Meanwhile, temporal heterogeneity arises because, over time, each client may collect new non-IID data that varies in domains and categories.
For example, over time, various hospitals around the world may encounter previously unseen diseases, such as the emergence of COVID-19 and its new variants.

The main negative consequence of spatio-temporal data heterogeneity in FCL is spatio-temporal catastrophic forgetting, as shown in Figure~\ref{fig:fig1} (b).
The direct cause of catastrophic forgetting is the overwriting of well-learned model parameters from previous tasks.
Let's consider that each client has trained a well-performing model on the respective local task.
When the server aggregates the local models, it can overwrite critical task-specific parameters, leading to spatial forgetting of local knowledge.
At the same time, temporal data heterogeneity in the online task stream can introduce task-recency biases in the model parameters, causing temporal forgetting of previous knowledge. 
Since clients use the aggregated global model as the basis to train on subsequent tasks, spatial and temporal forgetting tend to influence each other, exacerbating the challenges in FCL.

To better analyze the underlying causes of spatio-temporal forgetting in FCL, we illustrate the gradient updating for different non-IID scenarios in Figure~\ref{fig:Impact-gradient}.
In the ideal case of an IID setting, the gradient updates are expected to remain consistent and unified, as shown in Figure~\ref{fig:Impact-gradient} (a).
However, due to the inaccessibility of local data, spatial heterogeneity causes each local client to update the gradient in conflicting directions, leading to spatial forgetting, as shown in Figure~\ref{fig:Impact-gradient} (b).
In addtion, due to the unavailability of previous data, temporal heterogeneity causes the current gradient update to conflict with the historical gradient updates in each round, resulting in temporal forgetting, as shown in Figure~\ref{fig:Impact-gradient} (c).
In FCL, the simultaneous and intertwined occurrence of both temporal and spatial heterogeneity leads to more severe spatio-temporal forgetting, as shown in Figure~\ref{fig:Impact-gradient} (d).

Building on the above analysis, we identify that the root cause of spatio-temporal catastrophic forgetting lies in the inherent vulnerability and sensitivity of gradients to non-IID data.
Thus, we believe that to fully address the issue of spatio-temporal catastrophic forgetting, it is essential to seek methods that avoid gradient-based updates throughout the entire training process in FCL.

\section{Appendix for Theoretical Analysis on Lemma 1} 
\label{subsection:appendix.1}

\noindent\textbf{Lemma 1:} For an optimization problem of the following form:
\begin{equation}
    \label{eq:appendix-theorem-1-1}
    \operatorname*{argmin}_{\mathbf{W}} \; 
    {\left\|\mathbf{Y} -\mathbf{F} \mathbf{W} \right\|}_\mathrm{F}^2+\gamma{\left\| \mathbf{W} \right\|}_\mathrm{F}^2,
\end{equation}
there exists a closed-form solution given by:
\begin{equation}
    \label{eq:appendix-theorem-1-2}
    \mathbf{W}=
    {({\mathbf{F}}^\top\mathbf{F}+\gamma\mathbf{I})}^{-1}{\mathbf{F}}^\top\mathbf{Y}.
\end{equation}

\noindent \textbf{\textit{Proof.}} The objective function in \eqref{eq:appendix-theorem-1-1} can be rewritten as:
\begin{equation}
    \left\|\mathbf{Y}-\mathbf{F}\mathbf{W}\right\|_F^2 + \gamma \left\|\mathbf{W}\right\|_F^2
    =\mathrm{Tr}\left[(\mathbf{Y}-\mathbf{F}\mathbf{W})^\top(\mathbf{Y}-\mathbf{F}\mathbf{W})\right] + \gamma \mathrm{Tr}( \mathbf{W}^\top \mathbf{W} ),
\end{equation}
which can be further expanded as:
\begin{equation}
    \mathrm{Tr}(\mathbf{Y}^\top \mathbf{Y}) - 2\cdot\mathrm{Tr}(\mathbf{Y}^\top \mathbf{F} \mathbf{W}) + \mathrm{Tr}( \mathbf{W}^\top \mathbf{F}^\top \mathbf{F} \mathbf{W}) + \gamma \mathrm{Tr}(\mathbf{W}^\top \mathbf{W}).
\end{equation}

To determine the closed-form solution $\mathbf{\hat W}$, let's compute the derivative of the objective function with respect to $\mathbf{W}$ as follows:
\begin{equation}
\frac{\partial}{\partial \mathbf{W}} 
    \left( \left\| \mathbf{Y} - \mathbf{F} \mathbf{W} \right\|_F^2 + \gamma \left\| \mathbf{W} \right\|_\mathrm{F}^2 \right)
= -2\mathbf{F}^\top \mathbf{Y} + 2 \mathbf{F}^\top \mathbf{F} \mathbf{W} + 2 \gamma \mathbf{W}.
\end{equation}

By setting the derivative to zero, we can obtain:
\begin{equation}
\begin{aligned}
   ( \mathbf{F}^\top \mathbf{F} + \gamma\mathbf{I}) \mathbf{W} = \mathbf{F}^\top \mathbf{Y}.
\end{aligned}
\end{equation}
Since $(\mathbf{F}^\top \mathbf{F} + \gamma\mathbf{I})$ is positive-definite, we can obtain:
\begin{equation}
\mathbf{W} =  (\mathbf{F}^\top \mathbf{F} + \gamma\mathbf{I})^{-1} \mathbf{F}^\top \mathbf{Y}.
\end{equation}

\hfill $\blacksquare$

\clearpage

\section{Appendix for Theoretical Analysis on Theorem 1} 
\label{subsection:appendix.2}

Here, we theoretically prove that the global model recursively derived through our AFCL is exactly equivalent to the optimal global model obtained via centralized joint learning over the complete dataset.
Given that the \textit{Global Knowledge Matrix} serves as the foundation for computing the global model in our AFCL, we first establish the validity of the recursive update method for the \textit{Global Knowledge Matrix} and provide its analytical expression in \textbf{Lemma 2}.

\noindent\textbf{Lemma 2:} We consider the following recursive update formulation between $\mathbf{G}_{k-1}$ and $\mathbf{G}_{k}$:
\begin{equation}
\label{eq:appendix-lemma-1-1}
\mathbf{G}_{k} = \begin{bmatrix} \mathbf{A}_{k}  \mathbf{G}_{k-1} + \mathbf{B}_k  \mathbf{\hat W}_k & \mathbf{B}_k  \mathbf{\check W}_k \end{bmatrix},
k\in \left(1,K \right],
\end{equation}
where
\begin{equation}
\label{eq:appendix-lemma-1-2}
\begin{cases}
\mathbf{A}_{k} = \mathbf{I} - (\mathbf{\tilde R}_{k-1})^{-1} \mathbf{R}_k (\mathbf{I} - (\mathbf{\tilde R}_{k})^{-1} \mathbf{R}_k), & \mathbf{\tilde R}_{k} = \sum_{i=1}^k \mathbf{R}_i, \\
\mathbf{B}_{k} = \mathbf{I} - (\mathbf{R}_k)^{-1} \mathbf{\tilde R}_{k-1} (\mathbf{I} - (\mathbf{\tilde R}_{k})^{-1} \mathbf{\tilde R}_{k-1}),  & \mathbf{R}_i = \mathbf{F}_i^\top \mathbf{F}_i +\gamma\mathbf{I}.
\end{cases}
\end{equation}
The closed-form expression of $\mathbf{G}_{k}$ is given by:
\begin{equation}
\label{eq:appendix-lemma-1-3}
\mathbf{G}_{k} = (\mathbf{F}_{1:k}^\top\mathbf{F}_{1:k} + k \gamma \mathbf{I}) ^{-1} ( \mathbf{F}_{1:k}^\top\mathbf{Y}_{1:k} ).
\end{equation}

\noindent \textbf{\textit{Proof.}} 
We employ mathematical induction to prove the theorem in detail. 
Specifically, we first establish the base case, i.e., for the first joined (virtual) client, the initial \textit{Global Knowledge Matrix} $\mathbf{G}_{1}$ satisfies \eqref{eq:appendix-lemma-1-3}.
Then, assuming that \eqref{eq:appendix-lemma-1-3} holds for the arbitrary $\mathbf{G}_k$, we prove that the corresponding $\mathbf{G}_{k+1}$ obtained via \eqref{eq:appendix-lemma-1-1} and \eqref{eq:appendix-lemma-1-2}, also satisfies \eqref{eq:appendix-lemma-1-3}.
The detailed proof is presented as follows:

\textbf{(1) Base Case:}
Let's consider the initial case for the first joined (virtual) client.
As detailed in Section \ref{subsection:3.4}, the initial value of the \textit{Global Knowledge Matrix} is set to $\mathbf{G}_1 = \mathbf{\check W}_1$.
Since no local model exists for known classes at this point, and based on \eqref{eq:theorem-1-2}, $\mathbf{G}_1$ can be numerically expressed as: 
\begin{equation}
\label{eq:appendix-lemma-1-4}
\mathbf{G}_1 = \mathbf{\check W}_1 = (\mathbf{F}_1^\top\mathbf{F}_1 + \gamma \mathbf{I}) ^{-1} ( \mathbf{F}_1^\top\mathbf{Y}_1 ),
\end{equation}
which clearly satisfies the form in \eqref{eq:appendix-lemma-1-3}.

\textbf{(2) Inductive Hypothesis:}
Assume that the \textit{Global Knowledge Matrix} $\mathbf{G}_k$ obtained after aggregating the first $k$ clients satisfies \eqref{eq:appendix-lemma-1-3}, i.e.,
\begin{equation}
\label{eq:appendix-lemma-1-5}
\mathbf{G}_k = (\mathbf{F}_{1:k}^\top\mathbf{F}_{1:k} + k \gamma \mathbf{I}) ^{-1} ( \mathbf{F}_{1:k}^\top\mathbf{Y}_{1:k} ).
\end{equation}

\textbf{(3) Inductive Step:}
Here, we further demonstrate that $\mathbf{G}_{k+1}$, derived from $\mathbf{G}_k$ using \eqref{eq:appendix-lemma-1-1} and \eqref{eq:appendix-lemma-1-2}, satisfies \eqref{eq:appendix-lemma-1-3}.
By substituting\eqref{eq:appendix-lemma-1-3} into $\mathbf{A}_{k+1}  \mathbf{G}_{k}$, we can obtain:
\begin{equation}
\label{eq:appendix-lemma-1-6}
  \begin{aligned}
\mathbf{A}_{k+1}  \mathbf{G}_{k} &= [ \mathbf{I} - (\mathbf{\tilde R}_{k})^{-1} \mathbf{R}_{k+1} (\mathbf{I} - (\mathbf{\tilde R}_{k+1})^{-1} \mathbf{R}_{k+1}) ] (\mathbf{\tilde R}_{k})^{-1} \mathbf{F}_{1:k}^\top\mathbf{Y}_{1:k} \\
&= [ \mathbf{I} - (\mathbf{\tilde R}_{k})^{-1} \mathbf{R}_{k+1} (\mathbf{I} - (\mathbf{R}_{k+1} + \mathbf{\tilde R}_{k})^{-1} \mathbf{R}_{k+1}) ] (\mathbf{\tilde R}_{k})^{-1} \mathbf{F}_{1:k}^\top\mathbf{Y}_{1:k} \\
&= [ \mathbf{I} - (\mathbf{\tilde R}_{k})^{-1}  (\mathbf{R}_{k+1} - \mathbf{R}_{k+1}(\mathbf{R}_{k+1} + \mathbf{\tilde R}_{k})^{-1} \mathbf{R}_{k+1}) ] (\mathbf{\tilde R}_{k})^{-1} \mathbf{F}_{1:k}^\top\mathbf{Y}_{1:k}.
  \end{aligned}
\end{equation}

According to the Woodbury Matrix Identity, for any matrices $\mathbf{A}\in\mathbb{R}^{n\times n}$, $\mathbf{B}\in\mathbb{R}^{n\times m}$, $\mathbf{C}\in\mathbb{R}^{m\times m}$, and $\mathbf{D}\in\mathbb{R}^{m\times n}$, the following equation holds consistently:
\begin{equation}
\label{eq:appendix-lemma-1-7}
(\mathbf{A}+\mathbf{B}\mathbf{C}\mathbf{D})^{-1}=\mathbf{A}^{-1}-\mathbf{A}^{-1}\mathbf{B}(\mathbf{C}^{-1}+\mathbf{D}\mathbf{A}^{-1}\mathbf{B})^{-1}\mathbf{D}\mathbf{A}^{-1}.
\end{equation}

Subsequently, based on \eqref{eq:appendix-lemma-1-7}, we substitute both $\mathbf{B}$ and $\mathbf{D}$ with $\mathbf{I}$. This adjustment allows us to derive:
\begin{equation}
\label{eq:appendix-lemma-1-8}
(\mathbf{A}+\mathbf{C})^{-1}=\mathbf{A}^{-1}-\mathbf{A}^{-1}(\mathbf{A}^{-1}+\mathbf{C}^{-1})^{-1}\mathbf{A}^{-1}.
\end{equation}

Meanwhile, we can further obtain:
\begin{equation}
\label{eq:appendix-lemma-1-9}
\mathbf{A}^{-1}(\mathbf{A}^{-1}+\mathbf{C}^{-1})^{-1}\mathbf{A}^{-1} = \mathbf{A}^{-1} - (\mathbf{A}+\mathbf{C})^{-1}.
\end{equation}

By replacing $\mathbf{A}^{-1}$ and $\mathbf{C}^{-1}$ in \eqref{eq:appendix-lemma-1-9} with $\mathbf{R}_{k+1}$ and $\mathbf{\tilde R}_{k}$ respectively, yields:
\begin{equation}
\label{eq:appendix-lemma-1-10}
\mathbf{R}_{k+1}(\mathbf{R}_{k+1}+\mathbf{\tilde R}_{k})^{-1}\mathbf{R}_{k+1} = \mathbf{R}_{k+1} - ( (\mathbf{\tilde R}_{k})^{-1} + (\mathbf{R}_{k+1})^{-1} )^{-1}.
\end{equation}

Incorporating \eqref{eq:appendix-lemma-1-10} into \eqref{eq:appendix-lemma-1-6}, we can find:
\begin{equation}
\label{eq:appendix-lemma-1-11}
  \begin{aligned}
\mathbf{A}_{k+1}  \mathbf{G}_{k} &= [ \mathbf{I} - (\mathbf{\tilde R}_{k})^{-1}  (\mathbf{R}_{k+1} - \mathbf{R}_{k+1}(\mathbf{R}_{k+1} + \mathbf{\tilde R}_{k})^{-1} \mathbf{R}_{k+1}) ] (\mathbf{\tilde R}_{k})^{-1} \mathbf{F}_{1:k}^\top\mathbf{Y}_{1:k}\\
&= [ \mathbf{I} - (\mathbf{\tilde R}_{k})^{-1}  (\mathbf{R}_{k+1} - \mathbf{R}_{k+1} + ( (\mathbf{\tilde R}_{k})^{-1} + (\mathbf{R}_{k+1})^{-1} )^{-1} )] (\mathbf{\tilde R}_{k})^{-1} \mathbf{F}_{1:k}^\top\mathbf{Y}_{1:k} \\
&= [ \mathbf{I} - (\mathbf{\tilde R}_{k})^{-1}  ( (\mathbf{\tilde R}_{k})^{-1} + (\mathbf{R}_{k+1})^{-1} )^{-1} ] (\mathbf{\tilde R}_{k})^{-1} \mathbf{F}_{1:k}^\top\mathbf{Y}_{1:k}.
  \end{aligned}
\end{equation}

\noindent Subsequently, replacing $\mathbf{A}$ and $\mathbf{C}$ in \eqref{eq:appendix-lemma-1-9} with $\mathbf{\tilde R}_{k}$ and $\mathbf{R}_{k+1}$ respectively, yields:
\begin{equation}
\label{eq:appendix-lemma-1-12}
(\mathbf{\tilde R}_{k})^{-1} ( (\mathbf{\tilde R}_{k})^{-1} + (\mathbf{R}_{k+1})^{-1} )^{-1} (\mathbf{\tilde R}_{k})^{-1} = (\mathbf{\tilde R}_{k})^{-1} - (\mathbf{\tilde R}_{k}+\mathbf{R}_{k+1})^{-1}.
\end{equation}

By incorporating \eqref{eq:appendix-lemma-1-12} into \eqref{eq:appendix-lemma-1-11}, we can obtain:
\begin{equation}
\label{eq:appendix-lemma-1-13}
  \begin{aligned}
\mathbf{A}_{k+1}  \mathbf{G}_{k} &= [ \mathbf{I} - (\mathbf{\tilde R}_{k})^{-1}  ( (\mathbf{\tilde R}_{k})^{-1} + (\mathbf{R}_{k+1})^{-1} )^{-1} ] (\mathbf{\tilde R}_{k})^{-1} \mathbf{F}_{1:k}^\top\mathbf{Y}_{1:k} \\
&= [ (\mathbf{\tilde R}_{k})^{-1} - (\mathbf{\tilde R}_{k})^{-1}  ( (\mathbf{\tilde R}_{k})^{-1} + (\mathbf{R}_{k+1})^{-1} )^{-1} (\mathbf{\tilde R}_{k})^{-1}]  \mathbf{F}_{1:k}^\top\mathbf{Y}_{1:k} \\
&= [ (\mathbf{\tilde R}_{k})^{-1} - (\mathbf{\tilde R}_{k})^{-1} + (\mathbf{\tilde R}_{k}+\mathbf{R}_{k+1})^{-1}]  \mathbf{F}_{1:k}^\top\mathbf{Y}_{1:k} \\
&= (\mathbf{\tilde R}_{k+1})^{-1}  \mathbf{F}_{1:k}^\top\mathbf{Y}_{1:k}.
  \end{aligned}
\end{equation}

Similarly, we can obtain:
\begin{equation}
\label{eq:appendix-lemma-1-14}
\mathbf{B}_{k+1}  \mathbf{\hat W}_{k+1}= (\mathbf{\tilde R}_{k+1})^{-1} \mathbf{F}_{k+1}^\top \mathbf{\hat Y}_{k+1},
\end{equation}
\begin{equation}
\label{eq:appendix-lemma-1-15}
\mathbf{B}_{k+1}  \mathbf{\check W}_{k+1} = (\mathbf{\tilde R}_{k+1})^{-1} \mathbf{F}_{k+1}^\top \mathbf{\check Y}_{k+1}.
\end{equation}

By substituting \eqref{eq:appendix-lemma-1-13}, \eqref{eq:appendix-lemma-1-14}, and \eqref{eq:appendix-lemma-1-15} into \eqref{eq:appendix-lemma-1-1}, we can further derive:
\begin{equation}
\label{eq:appendix-lemma-1-16}
\begin{aligned}
    \mathbf{G}_{k+1} &= \begin{bmatrix} \mathbf{A}_{k+1}  \mathbf{G}_{k} + \mathbf{B}_{k+1}  \mathbf{\hat W}_{k+1} & \mathbf{B}_{k+1}  \mathbf{\check W}_{k+1} \end{bmatrix} \\
    &= \begin{bmatrix} (\mathbf{\tilde R}_{k+1})^{-1}  \mathbf{F}_{1:k}^\top\mathbf{Y}_{1:k} + (\mathbf{\tilde R}_{k+1})^{-1} \mathbf{F}_{k+1}^\top \mathbf{\hat Y}_{k+1} & (\mathbf{\tilde R}_{k+1})^{-1} \mathbf{F}_{k+1}^\top \mathbf{\check Y}_{k+1} \end{bmatrix} \\
    &= (\mathbf{\tilde R}_{k+1})^{-1} \begin{bmatrix} \mathbf{F}_{1:k}^\top\mathbf{Y}_{1:k} + \mathbf{F}_{k+1}^\top \mathbf{\hat Y}_{k+1} & \mathbf{F}_{k+1}^\top \mathbf{\check Y}_{k+1} \end{bmatrix} \\
    &=\begin{bmatrix} \sum_{i=1}^{k+1} ( \mathbf{F}_i^\top\mathbf{F}_i + \gamma \mathbf{I})\end{bmatrix}^{-1} \begin{bmatrix}\mathbf{F}_{1:k}^\top\mathbf{Y}_{1:k} + \mathbf{F}_{k+1}^\top\mathbf{\hat Y}_{k+1}  & \mathbf{F}_{k+1}^\top\mathbf{\check Y}_{k+1}   \end{bmatrix}.
\end{aligned}
\end{equation}

Considering the presence of known and unknown classes in a class-incremental scenario, the complete feature matrix $\mathbf{F}_{1:k+1}$ and the corresponding label matrix $\mathbf{Y}_{1:k+1}$ satisfy:
\begin{equation}
\label{eq:appendix-lemma-1-17}
\mathbf{F}_{1:k+1} = \begin{bmatrix} \mathbf{F}_{1:k}\\ \mathbf{F}_{k+1} \end{bmatrix}, \quad \mathbf{Y}_{1:k+1} = \begin{bmatrix} \mathbf{Y}_{1:k} & \mathbf{0}\\ \mathbf{\hat Y}_{k+1} & \mathbf{\check Y}_{k+1} \end{bmatrix}.
\end{equation}

Therefore, we can obtain:
\begin{equation}
\label{eq:appendix-lemma-1-18}
\begin{aligned}
    \mathbf{F}_{1:k+1}^\top \mathbf{F}_{1:k+1} + (k+1) \gamma \mathbf{I} &= \begin{bmatrix} \mathbf{F}_{1:k}^\top & \mathbf{F}_{k+1}^\top \end{bmatrix} \begin{bmatrix} \mathbf{F}_{1:k}\\ \mathbf{F}_{k+1} \end{bmatrix}  + (k+1) \gamma \mathbf{I}\\
    &= \mathbf{F}_{1:k}^\top \mathbf{F}_{1:k} + \mathbf{F}_{k+1}^\top \mathbf{F}_{k+1}  + (k+1) \gamma \mathbf{I} \\
    &= \sum_{i=1}^{k+1} ( \mathbf{F}_i^\top\mathbf{F}_i + \gamma \mathbf{I}),
\end{aligned}
\end{equation}
\begin{equation}
\label{eq:appendix-lemma-1-19}
\begin{aligned}
    \mathbf{F}_{1:k+1}^\top \mathbf{Y}_{1:k+1} &= \begin{bmatrix} \mathbf{F}_{1:k}^\top & \mathbf{F}_{k+1}^\top \end{bmatrix} \begin{bmatrix} \mathbf{Y}_{1:k} & \mathbf{0}\\ \mathbf{\hat Y}_{k+1} & \mathbf{\check Y}_{k+1} \end{bmatrix} \\
    &= \begin{bmatrix} \mathbf{F}_{1:k}^\top \mathbf{Y}_{1:k} + \mathbf{F}_{k+1}^\top \mathbf{\hat Y}_{k+1} & \mathbf{F}_{k+1}^\top \mathbf{\check Y}_{k+1} \end{bmatrix}.
\end{aligned}
\end{equation}

By substituting \eqref{eq:appendix-lemma-1-18} and \eqref{eq:appendix-lemma-1-19} into \eqref{eq:appendix-lemma-1-16}, we can further derive:
\begin{equation}
\label{eq:appendix-lemma-1-20}
\begin{aligned}
    \mathbf{G}_{k+1} &= \begin{bmatrix} \sum_{i=1}^{k+1} ( \mathbf{F}_i^\top\mathbf{F}_i + \gamma \mathbf{I})\end{bmatrix}^{-1} \begin{bmatrix}\mathbf{F}_{1:k}^\top\mathbf{Y}_{1:k} + \mathbf{F}_{k+1}^\top\mathbf{\hat Y}_{k+1}  & \mathbf{F}_{k+1}^\top\mathbf{\check Y}_{k+1}   \end{bmatrix} \\
    &= \begin{bmatrix}  \mathbf{F}_{1:k+1}^\top \mathbf{F}_{1:k+1} + (k+1) \gamma \mathbf{I} \end{bmatrix}^{-1} (\mathbf{F}_{1:k+1}^\top \mathbf{Y}_{1:k+1}).
\end{aligned}
\end{equation}

The result shows that if $\mathbf{G}_k$ satisfies \eqref{eq:appendix-lemma-1-3}, then the corresponding $\mathbf{G}_{k+1}$ obtained from \eqref{eq:appendix-lemma-1-1} and \eqref{eq:appendix-lemma-1-2} also satisfies \eqref{eq:appendix-lemma-1-3}.

\textbf{(4) Conclusion:} 
By the principle of mathematical induction, for any \textit{Global Knowledge Matrix} $\mathbf{G}_k$, the corresponding closed-form expression is given by \eqref{eq:appendix-lemma-1-3}.
This also confirms the validity of our recursive update method for the \textit{Global Knowledge Matrix}.

\hfill $\blacksquare$

\clearpage

Based on \textbf{Lemma 2}, we can obtain the analytical expression of the \textit{Global Knowledge Matrix}. 
Building on this result, we further analyze the validity of our AFCL and theoretically prove that the global model recursively derived through \eqref{eq:subsection-3.5-4} is exactly equivalent to the optimal global model obtained via centralized joint learning over the complete dataset in \textbf{Theorem 1}.

\noindent\textbf{Theorem 1:} We consider the following computational formulation:
\begin{equation}
\label{eq:appendix-theorem-2-1}
  \begin{aligned}
\mathbf{W}_k=[\mathbf{\tilde R}_{k} - (k-1)\gamma \mathbf{I}]^{-1} \mathbf{\tilde R}_{k} \mathbf{G}_{k},
\end{aligned}
\end{equation}
where $\mathbf{G}_{k}$ and $\mathbf{\tilde R}_{k}$ are obtained from \eqref{eq:subsection-3.5-2:GKM update} and \eqref{eq:subsection-3.5-3}, respectively.
The computation result of \eqref{eq:appendix-theorem-2-1} is exactly equivalent to \eqref{eq:theorem-1-3}, which corresponds to the optimal solution (i.e., the centralized joint learning) of empirical risk minimization over the full datasets $\mathcal{D}_{1:k}$ from the first $k$ clients.

\noindent \textbf{\textit{Proof.}}
By further transforming the analytical expression of $\mathbf{G}_{k}$ derived in \textbf{Lemma 2}, we can obtain:
\begin{equation}
\label{eq:appendix-theorem-2-2}
\begin{aligned}
    \mathbf{G}_{k} &= (\mathbf{F}_{1:k}^\top\mathbf{F}_{1:k} + k \gamma \mathbf{I}) ^{-1} ( \mathbf{F}_{1:k}^\top\mathbf{Y}_{1:k} ) \\
    &= ( \sum\nolimits_{i=1}^k {\mathbf{F}}^\top_k {\mathbf{F}}_k + k \gamma \mathbf{I} )^{-1} ( \mathbf{F}_{1:k}^\top\mathbf{Y}_{1:k} ) \\
    &= \begin{bmatrix} \sum\nolimits_{i=1}^k ({\mathbf{F}}^\top_k {\mathbf{F}}_k + \gamma \mathbf{I}) \end{bmatrix}^{-1} ( \mathbf{F}_{1:k}^\top\mathbf{Y}_{1:k} ) \\
    &= ( \mathbf{\tilde R}_k ) ^{-1} ({\mathbf{F}}^\top_{1:k} \mathbf{Y}_{1:k}).
\end{aligned}
\end{equation}

By substituting \eqref{eq:appendix-theorem-2-2} into \eqref{eq:appendix-theorem-2-1}, we can further derive:
\begin{equation}
    \label{eq:appendix-theorem-2-3}
    \begin{aligned}
    \mathbf{W}_k
    &=[\mathbf{\tilde R}_{k} - (k-1)\gamma \mathbf{I}]^{-1} \mathbf{\tilde R}_{k} \mathbf{G}_{k} \\
    &=[\mathbf{\tilde R}_{k} - (k-1)\gamma \mathbf{I}]^{-1} \mathbf{\tilde R}_{k} ( \mathbf{\tilde R}_k ) ^{-1} ({\mathbf{F}}^\top_{1:k} \mathbf{Y}_{1:k}) \\
    &=( \sum\nolimits_{i=1}^k {\mathbf{F}}^\top_k  \mathbf{F}_k + \gamma\mathbf{I}) ^{-1} ({\mathbf{F}}^\top_{1:k} \mathbf{Y}_{1:k}) \\
    &={({\mathbf{F}}^\top_{1:k}  \mathbf{F}_{1:k}+\gamma\mathbf{I})}^{-1}({\mathbf{F}}^\top_{1:k} \mathbf{Y}_{1:k}).
    \end{aligned}
\end{equation}

It is evident that the computation result of \eqref{eq:appendix-theorem-2-1} is exactly equivalent to the optimal global model derived from \eqref{eq:theorem-1-3}, thereby validating the validity of our AFCL.

\section{Appendix for Theoretical Analysis on Theorem 2} 
\label{subsection:appendix.3}

Here, by leveraging the previously established \textbf{Lemma 1} and \textbf{Theorem 1}, we provide a detailed proof of \textbf{Theorem 2}, thereby theoretically validating the \textit{spatio-temporal invariance for non-IID data} and the \textit{order invariance across local clients} inherent in our AFCL.

\noindent\textbf{Theorem 2:} The final global model obtained by our AFCL is independent of the spatio-temporal data heterogeneity and client registration order, being identical to that obtained by centralized joint learning over the full dataset $\mathcal{D}_{1:K}$.

\noindent \textbf{\textit{Proof.}} 
First of all, let's analyze the fundamental impact of different cases of spatio-temporal data heterogeneity and client registration order, as outlined below:

\textbf{(1) Spatio-temporal data heterogeneity} reflects the non-IID characteristics of local datasets across (virtual) clients.
Therefore, different cases of spatio-temporal data heterogeneity fundamentally correspond to the reassignment of a subset of samples among clients' local datasets $\{\mathcal{D}_k\}_{k=1}^K$.

\textbf{(2) Client registration order} determines the sequencing of clients' local datasets within the complete dataset.
Therefore, different cases of client registration order fundamentally correspond to reordering the clients’ datasets $\{\mathcal{D}_k\}_{k=1}^K$ within the complete dataset $\mathcal{D}_{1:K}$.

In summary, different cases of spatio-temporal data heterogeneity and client registration order can be interpreted as reordering the samples within the complete dataset $\mathcal{D}_{1:K}$.

Subsequently, we further demonstrate that reordering the samples does not affect either the optimal model or the model obtained through our AFCL.
Specifically, we denote the reordered feature matrix and label matrix as $\mathbf{F}^*_{1:K}$ and $\mathbf{Y}^*_{1:K}$, respectively, which can be represented as:
\begin{equation}
\label{eq:appendix-theorem-3-1}
    \mathbf{F}^ *_{1:K} = \mathbf{P}\mathbf{F}_{1:K}, \quad \mathbf{Y}^ *_{1:K} = \mathbf{P}\mathbf{Y}_{1:K},
\end{equation}
where $\mathbf{P} \in \mathbb{R}^{N \times N}$ is the corresponding permutation matrix and satisfies $\mathbf{P}^{-1} = \mathbf{P}^\top$.
Meanwhile, we denote the optimal solution obtained using the reordered feature matrix and label matrix as $\mathbf{F}^*_{1:K}$ and $\mathbf{Y}^*_{1:K}$ as $\mathbf{\tilde W}_{K}^*$.
According to \eqref{eq:theorem-1-3}, $\mathbf{\tilde W}_{K}^*$ can be calculated as:
\begin{equation}
\label{eq:appendix-theorem-3-2}
    \mathbf{\tilde W}^*_{K} = {( {\mathbf{F}^*_{1:K}}^\top \mathbf{F}^*_{1:K} + \gamma\mathbf{I} )}^{-1} {\mathbf{F}^*_{1:K}}^\top \mathbf{Y}^*_{1:K}.
\end{equation}

By substituting \eqref{eq:appendix-theorem-3-1} into \eqref{eq:appendix-theorem-3-2}, we can further obtain:
\begin{equation}
\begin{aligned}
\label{eq:appendix-theorem-3-3}
    \mathbf{\tilde W} ^ *_K &= 
    {( {\mathbf{F}^*_{1:K}}^\top \mathbf{F}^*_{1:K} + \gamma\mathbf{I} )}^{-1} {\mathbf{F}^*_{1:K}}^\top \mathbf{Y}^*_{1:K}\\
    &= {( {\mathbf{F}_{1:K}}^\top \mathbf{P}^\top \mathbf{P} \mathbf{F}_{1:K} + \gamma\mathbf{I} )}^{-1} {\mathbf{F}_{1:K}}^\top \mathbf{P}^\top \mathbf{P} \mathbf{Y}_{1:K}\\
    &= {( {\mathbf{F}_{1:K}}^\top \mathbf{P}^{-1} \mathbf{P} \mathbf{F}_{1:K} + \gamma\mathbf{I} )}^{-1} {\mathbf{F}_{1:K}}^\top \mathbf{P}^\top \mathbf{P} \mathbf{Y}_{1:K}\\
    &={({\mathbf{F}_{1:K}}^\top\mathbf{F}_{1:K}+\gamma\mathbf{I})}^{-1}{\mathbf{F}_{1:K}}^\top\mathbf{Y}_{1:K}.
\end{aligned}
\end{equation}
It is evident that the optimal global model using the complete dataset via \eqref{eq:theorem-1-3} is entirely independent of the spatio-temporal data heterogeneity and the client registration order.
Subsequently, according to \textbf{Theorem 1}, the global model obtained recursively and distributively through our AFCL is fully equivalent to the optimal result derived from centralized joint learning over the full dataset.
Denoting the corresponding global model obtained by AFCL as $\mathbf{W}_K^*$, and we further derive:
\begin{equation}
\mathbf{W}_K^* = \mathbf{\tilde W }^*_K = {({\mathbf{F}_{1:K}}^\top\mathbf{F}_{1:K}+\gamma\mathbf{I})}^{-1}{\mathbf{F}_{1:K}}^\top\mathbf{Y}_{1:K}.
\end{equation}

It is evident that the global model recursively derived by our AFCL is also independent of the spatio-temporal data heterogeneity and the client registration order, and is exactly equivalent to the optimal result obtained by centralized joint learning over the complete dataset $\mathcal{D}_{1:K}$.

\hfill $\blacksquare$


\section{Appendix for Theoretical Analysis on System Efficiency} 
\label{subsection:appendix.4}

Here, we provide a detailed analysis of the system efficiency of our AFCL.
Specifically, to comprehensively demonstrate its effectiveness, we analyze the computational overhead and communication overhead on both the client and server sides.
The detailed analytical process is outlined as follows:

\textbf{(1) The computational overhead:}
First, let's focus on the computational overhead on the client side. 
Specifically, in each round, the corresponding client needs to compute its \textit{Regularized Gram Matrix} ${\mathbf{R}}_k$ using \eqref{eq:subsection-3.4-8}. 
Then the client needs to further compute its local models $\hat{\mathbf{W}}_k$ and $\check{\mathbf{W}}_k$ using \eqref{eq:subsection-3.4-6} and \eqref{eq:subsection-3.4-7}, which can be further simplified as:
\begin{equation}
\label{eq:appendix-4-1}
\hat{\mathbf{W}}_k  =({\mathbf{R}}_k)^{-1}\mathbf{F}_k^\top\mathbf{\hat Y}_k, \quad
  \check{\mathbf{W}}_k  =({\mathbf{R}}_k)^{-1}\mathbf{F}_k^\top\mathbf{\check Y}_k.
\end{equation}
Since $\mathbf{F}_k \in \mathbb{R}^{N_k \times l_e} $, $\mathbf{\hat Y}_k \in \mathbb{R}^{N_k \times d_{k-1}} $, and $\mathbf{\check Y}_k \in \mathbb{R}^{N_k \times (d_k - d_{k-1})} $, the computational complexities of computing $\mathbf{R}_k$, $\mathbf{\hat W}_k$, and $\mathbf{\check W}_k$ are $O({l_e}^2 N_k)$, $O({l_e}^3+{l_e}^2 N_k+{l_e}N_kd_{k-1})$, and $O({l_e}^3+{l_e}^2 N_k+{l_e}N_k(d_{k}-d_{k-1}))$, respectively.
Second, we further analyze the computational overhead on the server side.
Specifically, in each round, the server needs to update the \textit{Global Knowledge Matrix} to obtain $\mathbf{G}_k$ through \eqref{eq:subsection-3.5-2:GKM update} and \eqref{eq:subsection-3.5-3}, and subsequently compute the corresponding global model using $\mathbf{G}_k$ and $\mathbf{R}_k$ through \eqref{eq:subsection-3.5-4}.
Since $\mathbf{R}_k \in \mathbb{R}^{l_e \times l_e}$, $\mathbf{G}_{k-1} \in \mathbb{R}^{l_e \times d_k}$, $\mathbf{\hat W}_k \in \mathbb{R}^{l_e \times d_{k-1}}$, and $\mathbf{\check W} \in \mathbb{R}^{l_e \in (d_k-d_k-1)}$, the computational complexities for computing $\mathbf{G}_k$ and $\mathbf{W}_k$ are both $O({l_e}^3 + {l_e}^2d_k)$.

\textbf{(2) The communication overhead:} 
Unlike existing FCL methods, our AFCL does not require the server to transmit the global model to clients.
Therefore, we only need to analyze the communication overhead on the client side.
Specifically, in each round, the corresponding client transmits its \textit{Regularized Gram Matrix} ${\mathbf{R}}_k$, local weight for the known classes $\mathbf{\hat W}_k$, and local weight for the unknown classes $\mathbf{\check W}_k$ to the server.
Since ${\mathbf{R}}_k \in \mathbb{R}^{l_e \times l_e} $, $\mathbf{\hat W}_k \in \mathbb{R}^{l_e \times d_{k-1}}$, and $\mathbf{\check W}_k\in \mathbb{R}^{l_e \times (d_k-d_{k-1})}$, the communication complexity for the client side is $O({l_e}^2+l_ed_k)$.

In summary, in each round, the total computational complexity is $O({l_e}^3+{l_e}^2 N_k+{l_e}N_kd_{k})$ on the client side, and $O({l_e}^3 + {l_e}^2d_k)$ on the server side.
Additionally, only the client needs to upload their information (i.e., ${\mathbf{R}}_k$, $\mathbf{\hat W}_k$, and $\mathbf{\check W}_k$), with a total communication complexity of $O({l_e}^2+l_ed_k)$.

\clearpage

\section{Appendix for Experimental Details}
\label{app:Experiment}

\subsection{Details on Datasets \& Setting}
\label{app:Dataset Setting}
Here, we detail the dataset partitioning strategy used in our experiments.
First, to simulate temporal data heterogeneity, we adopt the well-established Si-Blurry setting \cite{Si-Blurry} to partition the entire dataset into $5$ or $10$ tasks.
Specifically, in the Si-Blurry setting, all classes are categorized into two groups: disjoint classes (with data exclusive to a specific task) and blurry classes (with data distributed across multiple tasks). The proportion of disjoint classes among the total classes is denoted as $r_\text{D}$.
Subsequently, a proportion $r_\text{B}$ of the data from each blurry class is randomly reassigned to other blurry classes. 
Each task’s data consisted of a mix of disjoint and blurry classes, with parameters set as $r_\text{B}=10 \%$ and $r_\text{D}=50 \%$.
Second, to simulate spatial data heterogeneity, we employ the Dirichlet distribution \cite{Dirichlet} to divide each task's dataset among $5$ clients.
Specifically, the data allocated to each client has an equal number of samples and follows the Dirichlet distribution, where the degree of spatial data heterogeneity among clients is controlled by the Dirichlet parameter $\alpha$.
For CIFAR-100 and Tiny-ImageNet, we set $\alpha \in \{0.1, 0.2\}$.
Given the higher complexity of ImageNet-R, baselines perform poorly on this dataset, making meaningful comparisons difficult. 
Therefore, we use $\alpha \in \{0.5, 1.0\}$ to reduce spatial heterogeneity and enhance the baselines' performance.
Notably, this adjustment does not affect our AFCL due to our inspiring property of spatio-temporal invariance, but merely brings baseline performance closer to ours for more meaningful comparisons.
Moreover, this setup aligns with the experimental settings adopted in existing studies \cite{FCL-7, FCL-10-Lander, Dirichlet-1, Dirichlet-2, Dirichlet-3}. 
All of the experiments are conducted on Nvidia RTX 4090D GPUs with 15 vCPUs Intel(R) Xeon(R).

\subsection{Details on Evaluation Metrics}
\label{app:Model Training}

Here, we provide a detailed description of the metrics employed in our experiments.
For convenience, let $\mathcal{A}_{j}^i$ denote the accuracy of the global model in the $i$-round on the test set of the $j$-th task. 
Based on this notation, the metrics are defined in detail as follows:
\begin{itemize}
    \item \textbf{Average Accuracy} is used to evaluate the overall performance of the method, computed as the average accuracy of the current global model on the test sets of all previously learned tasks.
    Let the average accuracy of the global model in the $i$-round be denoted as $\mathcal{A}_i$, which is calculated as:
    \begin{equation}
        \mathcal{A}_i = \frac{1}{i} \sum_{j=1}^i\mathcal{A}_{j}^i.
    \end{equation}
    \item \textbf{Average Knowledge Retention} is used to evaluate the retention of previously learned knowledge as tasks progress, computed as the average ratio of accuracy on each previously learned task’s test set between the global model when that task is first learned and the current global model.
    Let the average knowledge retention of the global model in the $i$-round be denoted as $\mathcal{F}_i$, which is calculated as:
    \begin{equation}
        \mathcal{F}_i = \frac{1}{i-1} \sum_{j=1}^{i-1} (\mathcal{A}_{j}^j / \mathcal{A}_{j}^i).
    \end{equation}
    \item \textbf{Average Stability} is used to evaluate the ability to retain previous task knowledge, computed as the mean accuracy of each round’s global model on the test sets of all tasks learned before that round.
    Let the average stability in the $i$-round be denoted as $\mathcal{S}_i$, which is calculated as:
    \begin{equation}
        \mathcal{S}_i = \frac{1}{i-1}\sum_{k=2}^{i} (\frac{1}{k-1}\sum_{j=1}^{k-1} \mathcal{A}_{j}^{k}).
    \end{equation}
    \item \textbf{Average Plasticity} is used to evaluate the ability to acquire new task knowledge, computed as the mean accuracy of each round’s global model on the test set of the task learned in that round.
    Let the average plasticity in the $i$-round be denoted as $\mathcal{P}_i$, which is calculated as:
    \begin{equation}
        \mathcal{P}_i = \frac{1}{i} \sum_{j=1}^i \mathcal{A}_{j}^j.
    \end{equation}
    \item \textbf{Cumulative Runtime} is used to evaluate the efficiency, computed as the total runtime required to complete training across all tasks, including both client-side local training and server-side global aggregation.
\end{itemize}

\subsection{Details on Baselines}

Here, we provide a detailed description of the baselines used in our experiments, along with their corresponding settings. 
Specifically, each selected baseline is outlined as follows:
\begin{itemize}
    \item \textbf{Finetune} \cite{fedavg}: This method is widely employed as a baseline in existing methods, where each client sequentially and simply learns tasks, and their local models are subsequently aggregated using FedAvg.
    \item \textbf{FedLwF} \cite{LwF}: This method incorporates replay-free continual learning techniques into traditional federated learning frameworks, leveraging the introduced knowledge distillation loss to preserve knowledge of previous tasks and thereby mitigate catastrophic forgetting.
    \item \textbf{TARGET} \cite{FCL-7}: This method is specifically designed for FCL based on generative replay-based techniques, mitigating catastrophic forgetting by synthesizing data of previous tasks at the server and distributing it to clients for local training.
    \item \textbf{FedEwc} \cite{Ewc}: This method incorporates replay-free continual learning techniques into traditional federated learning frameworks, utilizing regularization-based approaches to penalize modifications of critical parameters, thereby mitigating catastrophic forgetting.
    \item \textbf{FedIcaRL} \cite{iCaRL}: This method incorporates replay-based continual learning techniques into traditional federated learning frameworks, mitigating catastrophic forgetting by maintaining representative historical samples and utilizing them for subsequent model training.
    \item \textbf{FedMGP} \cite{FCL-2}: This method is specifically designed for FCL using replay-free techniques, which employs a frozen backbone and introduces coarse-grained global prompts and fine-grained local prompts to model shared and personalized knowledge across clients. 
    Only the global prompts are aggregated to mitigate the adverse effects of data heterogeneity.
    \item \textbf{FedCBC} \cite{FCL-3}: This method is specifically designed for FCL based on generative replay-based techniques, mitigating catastrophic forgetting by constructing independent variational autoencoder-based classifiers for each class and utilizing them to generate pseudo-samples that rehearse previous knowledge.
\end{itemize}

Next, we elaborate on the experimental settings for these baselines.
Overall, all methods are evaluated under the consistent FCL setting.
Additionally, we adopt different training strategies (i.e., communication rounds per task and local epochs per communication round) tailored to each method's characteristics.
Specifically, for FedMGP, each task involves 5 communication rounds, and clients perform 5 local epochs per communication round. 
FedCBC conducts 300 local epochs on each client with 50 communication rounds per task. 
For all other baseline methods, each task involves 100 communication rounds, and clients perform 5 local epochs per communication round.

\subsection{Detailed Results on Stability-Plasticity Analysis}
\label{app:Stability-Plasticity}
Here, we present detailed experimental results on the stability and plasticity of all methods, where stability reflects the ability to retain learned knowledge, and plasticity measures the ability to acquire new knowledge.
Specifically, we report the average stability and average plasticity across tasks, as shown in Tables~\ref{table:Stability}.
It is evident that our AFCL achieves superior performance in terms of both stability and plasticity across all datasets.
Specifically, our AFCL's final average stability is higher than that of the best baseline, FedCBC, by 35.25\%, 47.29\%, and 32.17\% on different datasets, and its final average plasticity is higher by 25.63\%, 38.35\%, and 35.22\%, respectively.
Notably, although some baselines exhibit high initial average stability and plasticity, their performance deteriorates significantly as the number of tasks increases, indicative of catastrophic forgetting.
In contrast, owing to the inspiring property of spatio-temporal invariance, the average stability and plasticity of our AFCL not only remain stable but actually improve with the growth in tasks.
This demonstrates that our method not only avoids spatio-temporal catastrophic forgetting but also continuously acquires new knowledge and reinforces learned knowledge during the continual learning process, ultimately achieving promising performance.

\begin{table*}[t]
    \centering
    \renewcommand{\arraystretch}{1.2}
    \caption{
    Stability and Plasticity of Different Methods in Processing Different Tasks.
    }
    \label{table:Stability}
    \resizebox{\textwidth}{!}{
        \begin{NiceTabular}{ | c | c c c c c c c c c >{\columncolor{mygreen!15}} c |} 
        \toprule
        \textbf{characteristic} & \textbf{Dataset} & \textbf{Task} & \textbf{Finetune} & \textbf{FedLwF} & \textbf{TARGET} & \textbf{FedEwc} & \textbf{FedIcaRL} & \textbf{FedMGP} & \textbf{FedCBC} & \textbf{AFCL}  \\
        \cline{1-11}
        \multirow{27}{*}{\textbf{Stability}}&\multirow{9}{*}{CIFAR-100}  
        & $2$ &  25.04  & 33.20  & \underline{49.90}  & 39.30  & 20.11  & \textbf{51.12}  & 45.20  & $38.90$   \\
&& $3$ &  17.50  & 20.83  & 33.50  & 28.50  & 22.36  & \underline{42.50}  & 41.82  & $\textbf{43.70}$   \\
&& $4$ &  12.01  & 15.63  & 30.40  & 25.83  & 9.75  & 33.66  & \textbf{49.53}  & $\underline{46.32}$   \\
&& $5$ &  12.17  & 17.68  & 15.38  & 17.93  & 11.78  & {35.44}  & \underline{35.61}  & $\textbf{50.13}$   \\
&& $6$ &  9.43  & 11.50  & 19.88  & 19.00  & 6.22  & 34.23  & \underline{34.98}  & $\textbf{52.52}$   \\
&& $7$ &  3.15  & 9.91  & 18.28  & 14.97  & 1.75  & 23.73  & \underline{32.23}  & $\textbf{54.52}$   \\
&& $8$ &  5.45  & 3.19  & 15.56  & 14.34  & 5.16  & 23.90  & \underline{32.70}  & $\textbf{56.03}$   \\
&& $9$ &  4.12  & 5.92  & 11.43  & 12.15  & 3.49  & 23.56  & \underline{36.78}  & $\textbf{56.87}$   \\
&& $10$ &  2.43  & 3.80  & 10.29  & 11.81  & 1.24  & 19.54  & \underline{29.45}  & $\textbf{58.57}$   \\

        \cline{2-11}
        &\multirow{9}{*}{Tiny-ImageNet} & $2$ &  24.02  & \underline{37.59}  & \textbf{54.50}  & 13.50  & 25.67  & 32.51  & {35.67}  & ${24.64}$   \\
&& $3$ &  8.97  & \underline{29.15}  & 1.30  & 12.10  & 19.93  & 26.11  & \textbf{36.68}  & ${22.09}$   \\
&& $4$ &  3.47  & 20.01  & 1.20  & 6.05  & 14.34  & 27.39  & \underline{37.57}  & $\textbf{38.47}$   \\
&& $5$ &  5.95  & 18.90  & 0.17  & 3.83  & 6.47  & 27.31  & \underline{37.06}  & $\textbf{41.83}$   \\
&& $6$ &  0.00  & 9.58  & 0.30  & 4.15  & 7.19  & 21.94  & \underline{29.74}  & $\textbf{44.86}$   \\
&& $7$ &  0.00  & 2.80  & 0.56  & 3.40  & 4.85  & 17.58  & \underline{31.22}  & $\textbf{48.23}$   \\
&& $8$ &  2.68  & 3.92  & 0.57  & 3.22  & 4.98  & 16.05  & \underline{29.67}  & $\textbf{50.21}$   \\
&& $9$ &  2.36  & 1.78  & 0.49  & 2.99  & 1.16  & 18.46  & \underline{27.48}  & $\textbf{52.40}$   \\
&& $10$ &  0.00  & 2.29  & 0.55  & 2.93  & 3.48  & 16.67  & \underline{20.98}  & $\textbf{53.94}$   \\

        \cline{2-11}
        &\multirow{9}{*}{ImageNet-R} & $2$ &  \underline{30.13}  & 23.32  & \textbf{49.90}  & 20.30  & 22.64  & 24.88  & 28.97  & ${18.06}$   \\
&& $3$ &  0.04  & 20.13  & 19.70  & 10.90  & 14.36  & \underline{26.04}  & \textbf{26.47}  & ${20.58}$   \\
&& $4$ &  0.57  & 11.33  & 5.40  & 8.75  & 13.00  & 18.97  & \textbf{26.44}  & $\underline{26.18}$   \\
&& $5$ &  0.43  & 9.39  & 2.70  & 4.87  & 11.76  & 21.95  & \underline{25.06}  & $\textbf{28.13}$   \\
&& $6$ &  0.00  & 11.27  & 2.08  & 5.15  & 7.52  & 22.21  & \underline{25.40}  & $\textbf{30.05}$   \\
&& $7$ &  0.00  & 7.94  & 1.76  & 4.78  & 1.91  & 10.90  & \underline{24.93 } & $\textbf{31.92}$   \\
&& $8$ &  0.00  & 0.00  & 1.30  & 4.30  & 3.46  & 13.46  & \underline{25.44}  & $\textbf{33.89}$   \\
&& $9$ &  1.44  & 2.11  & 1.17  & 4.10  & 4.83  & 15.92  & \underline{22.36}  & $\textbf{35.26}$   \\
&& $10$ &  0.00  & 6.81  & 1.09  & 3.69  & 7.71  & \underline{22.90}  & 18.68  & $\textbf{36.32}$   \\

        \cline{1-11}
        \multirow{30}{*}{\textbf{Plasticity}}&\multirow{10}{*}{CIFAR-100} & $1$ &  36.80  & 43.21  & \textbf{87.00}  & 49.30  & 35.70  & \underline{52.05}  & {49.88}  & ${30.10}$   \\
&& $2$ &  36.36  & 33.65  & 40.30  & 22.80  & 32.83  & \textbf{50.04}  & \underline{44.34}  & ${38.23}$   \\
&& $3$ &  37.15  & 30.94  & 42.80  & 23.90  & 29.20  & 35.29  & \underline{43.05}  & $\textbf{43.21}$   \\
&& $4$ &  28.93  & 18.72  & 15.70  & 6.20  & 34.24  & 33.47  & \textbf{49.12}  & $\underline{45.94}$   \\
&& $5$ &  33.45  & 20.04  & \underline{47.40}  & 16.20  & 32.55  & 33.41  & 39.26  & $\textbf{49.92}$   \\
&& $6$ &  \underline{33.47}  & 20.94  & 16.00  & 8.40  & 31.77  & 28.75  & {32.43}  & $\textbf{52.33}$   \\
&& $7$ &  27.77  & 12.90  & 10.30  & 8.50  & 35.76  & 22.65  & \underline{39.31}  & $\textbf{54.39}$   \\
&& $8$ &  31.98  & 15.23  & 13.40  & 10.00  & 30.03  & 18.60  & \underline{34.44}  & $\textbf{55.94}$   \\
&& $9$ &  30.69  & 12.46  & \underline{35.00}  & 18.20  & 32.90  & 21.15  & 30.61  & $\textbf{56.75}$   \\
&& $10$ &  27.90  & 7.60  & 16.80  & 8.50  & 28.10  & 16.17  & \underline{29.02}  & $\textbf{58.56}$   \\

        \cline{2-11}
        &\multirow{10}{*}{Tiny-ImageNet}& $1$ &  24.02  & \underline{37.59}  & \textbf{54.50}  & 13.50  & 25.67  & 32.51  & 35.67  & ${24.64}$   \\
&& $2$ &  20.60  & \textbf{35.46}  & 19.20  & 12.10  & 22.45  & 30.83  & \underline{31.50}  & ${21.42}$   \\
&& $3$ &  16.12  & 31.66  & 26.30  & 6.05  & 10.90  & 24.40  & \underline{33.54}  & $\textbf{37.98}$   \\
&& $4$ &  22.89  & 21.24  & 24.20  & 3.83  & 7.85  & 24.79  & \underline{30.60}  & $\textbf{41.46}$   \\
&& $5$ &  16.82  & 19.19  & 20.90  & 4.15  & 7.88  & 19.08  & \underline{30.43}  & $\textbf{44.83}$   \\
&& $6$ &  24.37  & 19.00  & 22.60  & 3.40  & 7.48  & 16.01  & \underline{27.13}  & $\textbf{48.17}$   \\
&& $7$ &  15.44  & 18.15  & \underline{27.40}  & 3.22  & 3.43  & 11.42  & 25.53  & $\textbf{50.18}$   \\
&& $8$ &  11.31  & 13.82  & 18.00  & 2.99  & 1.09  & 14.25  & \underline{25.90}  & $\textbf{52.41}$   \\
&& $9$ &  16.15  & 12.04  & 22.40  & 2.93  & 3.16  & 10.92  & \underline{25.63}  & $\textbf{53.85}$   \\
&& $10$ &  13.20  & 9.20  & 21.20  & 2.53  & 1.86  & 10.10  & \underline{24.62}  & $\textbf{54.79}$   \\
        \cline{2-11}
        &\multirow{10}{*}{ImageNet-R} & $1$ &  \underline{30.13}  & 23.32  & \textbf{49.90}  & 20.30  & 22.64  & 24.88  & 28.97  & ${18.06}$   \\
&& $2$ &  \underline{25.72}  & 23.25  & 13.80  & 9.20  & 15.58  & 23.68  & \textbf{27.34}  & ${22.41}$   \\
&& $3$ &  \textbf{28.26}  & 20.50  & 8.20  & 3.60  & 9.98  & 21.88  & \underline{26.93}  & ${24.56}$   \\
&& $4$ &  20.20  & 14.74  & 11.20  & 8.10  & 5.20  & \underline{21.43}  & 20.25  & $\textbf{26.75}$   \\
&& $5$ &  \underline{25.03}  & 21.05  & 5.60  & 6.00  & 5.57  & 19.43  & 16.12  & $\textbf{29.38}$   \\
&& $6$ &  16.42  & 14.23  & 1.80  & 4.60  & 1.69  & 14.60  & \underline{19.80}  & $\textbf{31.28}$   \\
&& $7$ &  15.30  & 14.77  & 8.10  & 4.50  & 7.10  & 8.51  & \underline{20.70}  & $\textbf{33.22}$   \\
&& $8$ &  \underline{17.68}  & 12.86  & 3.20  & 2.20  & 5.53  & 8.48  & 14.14  & $\textbf{35.18}$   \\
&& $9$ &  \underline{20.00 } & 13.00  & 2.80  & 1.20  & 2.74  & 3.83  & 15.31  & $\textbf{36.55}$   \\
&& $10$ &  \underline{16.50}  & 10.40  & 1.50  & 5.70  & 1.78  & 9.33  & 14.14  & $\textbf{38.05}$   \\

        \bottomrule
        \end{NiceTabular}
     }
\end{table*}

\clearpage
\subsection{Detailed Results on Efficiency Analysis}
\label{app:Efficiency}

Here, we conducted a detailed efficiency analysis of all methods to demonstrate the high efficiency of our AFCL. 
As detailed in Table~\ref{table:Efficiency}, we report the cumulative runtime for all methods evaluated on different datasets. 
Evidently, our AFCL significantly outperforms all baselines in efficiency, reducing the total runtime by 95.12\% ($\times 19.35$ speedup), 90.99\% ($\times 11.10$ speedup), and 86.65\% ($\times 7.49$ speedup) compared to the fastest baseline, Finetune, on the three datasets, respectively.
Notably, among all baselines, those with higher efficiency (e.g., Finetune and FedLwF) exhibit poorer performance, whereas the better-performing baselines (e.g., FedCBC and FedMGP) have lower efficiency.
Specifically, compared to the best-performing baseline, FedCBC, our AFCL’s efficiency advantage is even more pronounced, achieving the runtime reductions of 98.11\% ($\times 52.80$ speedup), 97.59\% ($\times 41.51$ speedup), and 96.58\% ($\times 29.25$ speedup) across the three datasets, respectively.
Meanwhile, compared to the second-performing baseline, FedMGP, our AFCL’s efficiency advantage is even more pronounced, achieving the runtime reductions of 98.51\% ($\times 67.31$ speedup), 98.21\% ($\times 55.85$ speedup), and 97.21\% ($\times 35.78$ speedup) across the three datasets, respectively.
In summary, our AFCL significantly surpasses existing methods not only in performance but also in efficiency, further demonstrating our AFCL's superiority.

\begin{table*}[t]
    \centering
    \renewcommand{\arraystretch}{1.2}
    \caption{
    Cumulative Runtime (Seconds) of Different Methods in Processing Different Tasks.
    }
    \label{table:Efficiency}
    \resizebox{\textwidth}{!}{
        \begin{NiceTabular}{ l c c c c c c c c >{\columncolor{mygreen!15}} c >{\columncolor{myblue!15}} c} 
        \toprule
        \textbf{Dataset} & \textbf{Task} & \textbf{Finetune} & \textbf{FedLwF} & \textbf{TARGET} & \textbf{FedEwc} & \textbf{FedIcaRL} & \textbf{FedMGP} & \textbf{FedCBC} & \textbf{AFCL} & \textbf{Improve}\\
        \cline{1-11}
        \multirow{10}{*}{CIFAR-100}
        &  $1$  &  $\underline{368.11} $  & $428.73 0$ & $659.79 $ & $502.04 $ & $569.29 $ & $1570.71 $ & $1297.10 $ & $\textbf{22.96}$ & $93.76 \%$  \\
        & $2$ &  $\underline{773.43} $  & $898.39 $ & $1305.09 $ & $955.46 $ & $1133.57 $ & $3215.38 $ & $2635.48 $ & $\textbf{48.09 }$ & $93.78  \%$ \\
        & $3$ &  $\underline{1216.98} $  & $1330.22 $ & $1957.66 $ & $1452.77 $ & $1670.89 $ & $4841.62 $ & $3939.27 $ & $\textbf{72.10 }$ & $94.08 \%$   \\
        & $4$ &  $\underline{1673.25} $  & $1804.43 $ & $2619.04 $ & $1936.42 $ & $2200.87 $ & $6538.77 $ & $5063.72 $ & $\textbf{94.93 }$ & $94.33 \%$   \\
        & $5$ &  $\underline{2151.61} $  & $2258.86 $ & $3257.32 $ & $2413.53 $ & $2742.69 $ & $8051.85 $ & $6345.23 $ & $\textbf{119.46 }$ & $94.45 \%$   \\
        & $6$ &  $\underline{2649.71} $  & $2756.45 $ & $3883.65 $ & $2870.39 $ & $3295.98 $ & $9702.75 $ & $7606.98 $ & $\textbf{144.55 }$ & $94.54 \%$   \\
        & $7$ &  $\underline{3165.80} $  & $3192.41 $ & $4489.37 $ & $3369.53 $ & $3877.50 $ & $11207.67 $ & $8815.03 $ & $\textbf{168.67 }$ & $94.67 \%$   \\
        & $8$ &  $3718.91 $  & $\underline{3687.14} $ & $5110.44 $ & $3873.96 $ & $4446.09 $ & $12886.32 $ & $10177.41 $ & $\textbf{190.88 }$ & $94.82 \%$   \\
        & $9$ &  $4286.94 $  & $\underline{4181.18} $ & $5757.81 $ & $4336.41 $ & $5002.01 $ & $14569.03 $ & $11485.06 $ & $\textbf{216.66 }$ & $94.82 \%$  \\
        & $10$ &  $4895.36 $  & $\underline{4627.22} $ & $6439.94 $ & $4816.03 $ & $5539.38 $ & $16095.30 $ & $12625.15 $ & $\textbf{239.11 }$ & $94.83 \%$   \\

        \cline{1-11}
        \multirow{10}{*}{Tiny-ImageNet} 
        & $1$ &  $\underline{385.79} $  & $476.70 $ & $743.95 $ & $570.93 $ & $620.62 $ & $1993.38 $ & $1509.69 $ & $\textbf{34.37 }$ & $91.09 \%$  \\
        & $2$ &  $\underline{827.25} $  & $953.93 $ & $1452.02 $ & $1152.51 $ & $1275.33 $ & $3999.32 $ & $2958.30 $ & $\textbf{68.18 }$ &  $91.76 \%$  \\
        & $3$ &  $\underline{1208.97} $  & $1425.36 $ & $2172.82 $ & $1712.69 $ & $1944.65 $ & $5938.53 $ & $4348.69 $ & $\textbf{104.18 }$ &  $91.38 \%$  \\
        & $4$ &  $\underline{1594.97} $  & $1882.55 $ & $2906.57 $ & $2305.47 $ & $2578.55 $ & $8100.52 $ & $5888.71 $ & $\textbf{137.40 }$ &  $91.39 \%$  \\
        & $5$ &  $\underline{2006.84} $  & $2361.92 $ & $3648.45 $ & $2914.96 $ & $3213.28 $ & $10041.26 $ & $7482.76 $ & $\textbf{176.46 }$ &  $91.21 \%$  \\
        & $6$ &  $\underline{2394.43} $  & $2841.04 $ & $4388.61 $ & $3476.39 $ & $3858.16 $ & $12202.70 $ & $9059.31 $ & $\textbf{215.91 }$ & $90.98 \%$   \\
        & $7$ &  $\underline{2787.50} $  & $3284.40 $ & $5105.60 $ & $4055.55 $ & $4493.40 $ & $14098.30 $ & $10444.75 $ & $\textbf{252.56 }$ & $90.94 \%$  \\
        & $8$ &  $\underline{3200.99} $  & $3726.47 $ & $5832.91 $ & $4628.79 $ & $5143.97 $ & $16147.96 $ & $12025.53 $ & $\textbf{289.34 }$ & $90.96 \%$   \\
        & $9$ &  $\underline{3580.38} $  & $4195.58 $ & $6544.12 $ & $5239.48 $ & $5779.07 $ & $18229.53 $ & $13599.33 $ & $\textbf{325.50 }$ & $90.91 \%$   \\
        & $10$ &  $\underline{4024.37} $  & $4649.91 $ & $7270.49 $ & $5837.14 $ & $6405.85 $ & $20241.53 $ & $15044.59 $ & $\textbf{362.45 }$ & $90.99 \%$   \\

        \cline{1-11}
        \multirow{10}{*}{ImageNet-R}
        &  $1$  &  $\underline{472.17} $  & $539.53 $ & $904.41 $ & $793.96 $ & $843.73 $ & $2232.91 $ & $1871.54 $ & $\textbf{64.60 }$ &  $86.32  \%$ \\
        & $2$ &  $\underline{922.71} $  & $1111.11 $ & $1768.40 $ & $1573.95 $ & $1649.67 $ & $4479.00 $ & $3756.79 $ & $\textbf{126.66 }$ &  $86.27 \%$ \\
        & $3$ &  $\underline{1410.65} $  & $1695.13 $ & $2667.68 $ & $2355.98 $ & $2469.21 $ & $6838.30 $ & $5565.48 $ & $\textbf{192.66 }$ & $86.34  \%$  \\
        & $4$ &  $\underline{1862.48} $  & $2276.86 $ & $3566.78 $ & $3097.92 $ & $3268.97 $ & $9084.73 $ & $7392.00 $ & $\textbf{253.59 }$ &  $86.38  \%$ \\
        & $5$ &  $\underline{2355.67} $  & $2821.88 $ & $4441.83 $ & $3880.85 $ & $4080.09 $ & $11262.95 $ & $9322.83 $ & $\textbf{315.22  }$  & $86.62  \%$ \\
        & $6$ &  $\underline{2838.58} $  & $3410.71 $ & $5308.43 $ & $4640.61 $ & $4889.48 $ & $13510.27 $ & $11188.07 $ & $\textbf{377.86  }$ & $86.69  \%$  \\
        & $7$ &  $\underline{3333.23} $  & $3972.22 $ & $6163.10 $ & $5393.25 $ & $5713.04 $ & $15851.26 $ & $13133.25 $ & $\textbf{439.14 }$ & $86.83  \%$  \\
        & $8$ &  $\underline{3786.72} $  & $4525.12 $ & $7026.21 $ & $6183.52 $ & $6551.65 $ & $18132.44 $ & $14882.78 $ & $\textbf{505.82 }$ & $86.64 \%$  \\
        & $9$ &  $\underline{4283.16} $  & $5061.42 $ & $7926.69 $ & $6934.51 $ & $7398.77 $ & $20370.29 $ & $16818.35 $ & $\textbf{573.41 }$ & $86.61 \%$  \\
        & $10$ &  $\underline{4758.67} $  & $5637.28 $ & $8810.50 $ & $7679.43 $ & $8246.49 $ & $22740.16 $ & $18590.48 $ & $\textbf{635.51 }$ & $86.65 \%$  \\
        \bottomrule
        \end{NiceTabular}
     }
\end{table*}




\end{document}